\documentclass[runningheads]{llncs}

\usepackage{eccv}

\usepackage{eccvabbrv}

\usepackage{graphicx}
\usepackage{booktabs}
\usepackage{multirow}
\usepackage{array}
\usepackage{graphicx}
\usepackage{subcaption}
\usepackage{adjustbox}
\usepackage{wrapfig}
\usepackage[table]{xcolor}
\usepackage{array}
\newcolumntype{C}[1]{>{\centering\arraybackslash}p{#1}}
\usepackage[accsupp]{axessibility}  

\usepackage{hyperref}

\usepackage{orcidlink}

\begin{document}

\title{ILV: Iterative Latent Volumes for Fast and Accurate Sparse-View CT Reconstruction} 
\titlerunning{ILV: Iterative Latent Volumes for Sparse-View CT}

\author{
Seungryong Lee\inst{1} \quad \hspace{-0.2cm}
Woojeong Baek\inst{2} \quad \hspace{-0.2cm}
Joosang Lee\inst{1} \quad \hspace{-0.2cm}
Eunbyung Park\inst{2}$^{\dagger}$
}

\authorrunning{S. Lee et al.}

\institute{
Sungkyunkwan University \and
Yonsei University
}




\maketitle

\begin{abstract}
A long-term goal in CT imaging is to achieve fast and accurate 3D reconstruction from sparse-view projections, thereby reducing radiation exposure, lowering system cost, and enabling timely imaging in clinical workflows.
Recent feed-forward approaches have shown strong potential toward this overarching goal, yet their results still suffer from artifacts and loss of fine details.
In this work, we introduce Iterative Latent Volumes (ILV), a feed-forward framework that integrates data-driven priors with classical iterative reconstruction principles to overcome key limitations of prior feed-forward models in sparse-view CBCT reconstruction.
At its core, ILV constructs an explicit 3D latent volume that is repeatedly updated by conditioning on multi-view X-ray features and the learned anatomical prior, enabling the recovery of fine structural details beyond the reach of prior feed-forward models. 
In addition, we develop and incorporate several key architectural components, including an X-ray feature volume, group cross-attention, efficient self-attention, and view-wise feature aggregation, that efficiently realize its core latent volume refinement concept. 
Extensive experiments on a large-scale dataset of approximately 14,000 CT volumes demonstrate that ILV significantly outperforms existing feed-forward and optimization-based methods in both reconstruction quality and speed. These results show that ILV enables fast and accurate sparse-view CBCT reconstruction suitable for clinical use. The project page is available at: \url{https://sngryonglee.github.io/ILV/}.
\end{abstract}

Computed Tomography (CT) is a widely used medical imaging technique that non-invasively visualizes the internal anatomy of the human body. By reconstructing a three-dimensional X-ray attenuation field from multiple two-dimensional projections acquired at different angles, CT enables detailed inspection of diverse tissues, including organs, bones, and tumors. Since it provides precise structural information, CT has become indispensable in numerous clinical applications such as disease diagnosis, treatment planning, and surgical navigation~\cite{cormack1963representation, cormack1964representation, hounsfield1973computerized, kak2001principles, otomo2022computed}.

However, clinical CT systems typically require hundreds of projection views to achieve sufficient image quality, which inevitably leads to substantial radiation exposure for patients. In addition, the installation and maintenance of CT equipment 
\begin{wrapfigure}{r}{0.5\linewidth} 
  \centering
  \includegraphics[width=\linewidth]{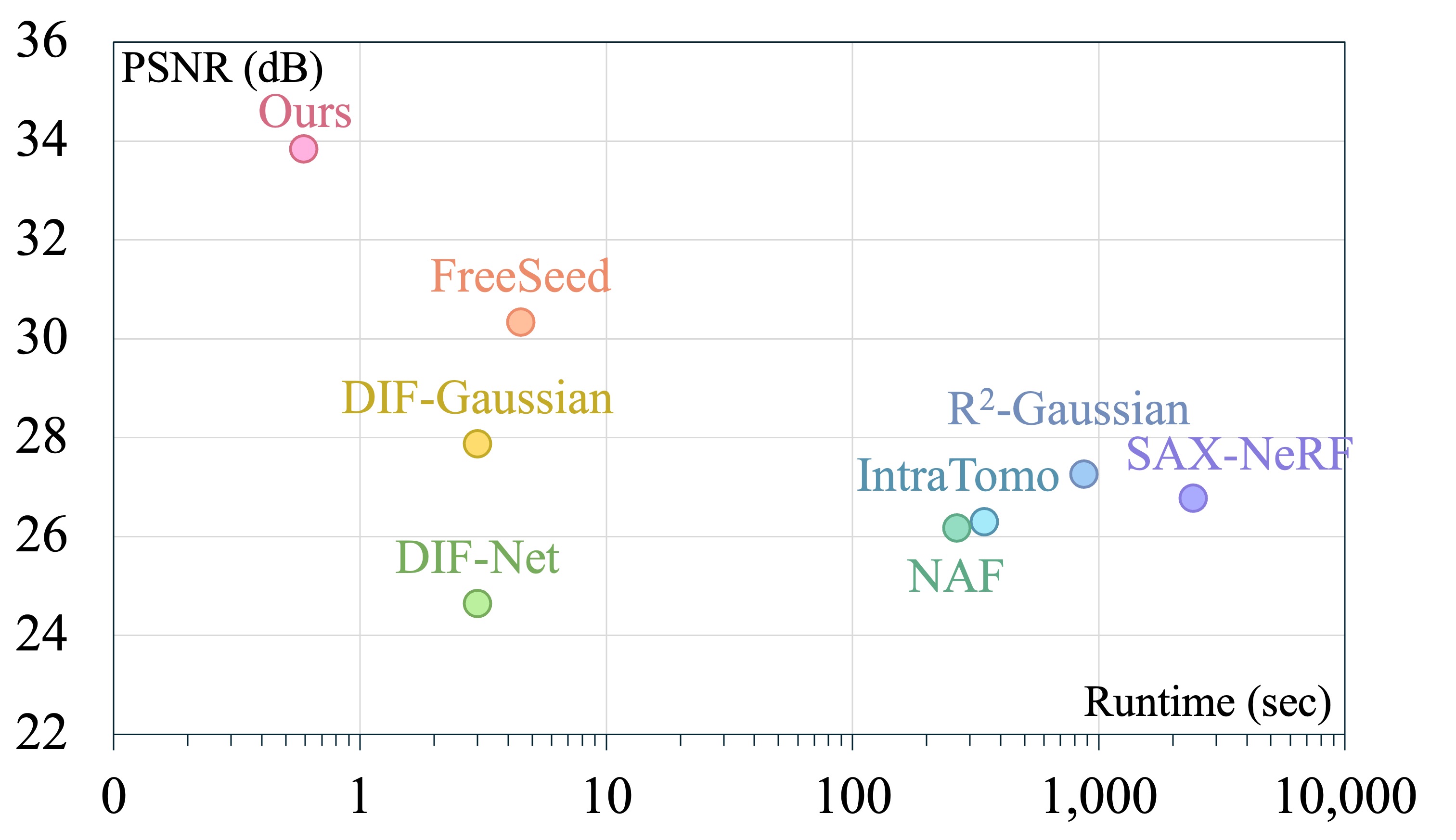}
  \caption{Comparison of the proposed method and state-of-the-art approaches in PSNR (dB) vs. Runtime (sec) plot. Runtime is presented on a logarithmic scale.}
  \label{fig1}
\end{wrapfigure}
demands costly hardware and supporting infrastructure, posing significant challenges in resource-limited clinical 
environments~\cite{hricak2021medical}. At the same time, rapid reconstruction has become increasingly important in time-critical applications, such as emergency triage or intraoperative guidance, where clinicians need near real-time imaging feedback. Consequently, the long-term goal of CT imaging is to enable high-quality reconstruction from sparse projections with minimal computational latency, a capability that remains challenging to achieve with existing methods.

Early studies employed analytical and iterative reconstruction algorithms~\cite{andersen1984simultaneous, feldkamp1984practical, pan2006variable, sidky2008image, yu2006region}. Analytical methods, such as FDK algorithm~\cite{feldkamp1984practical}, are fast and theoretically grounded but perform well only when projections are densely sampled, producing severe artifacts under sparse-view conditions. Iterative approaches, formulated as optimization problems that enforce data fidelity and prior regularization, alleviate these issues and yield improved performance in sparse-view settings, though their reconstruction quality remains limited. More recently, emerging 3D representations such as Neural Radiance Fields (NeRF)~\cite{mildenhall2021nerf} and 3D Gaussian Splatting (3DGS)~\cite{kerbl3Dgaussians} have shown strong potential for volumetric reconstruction and have been successfully applied to medical CT reconstruction~\cite{cai2024radiative, cai2024structure, zang2021intratomo, ruckert2022neat, zhaneural, zha2024r, shen2022nerp}. However, these methods are also based on time-consuming iterative optimization and exhibit limited performance under sparse-view conditions.

To achieve reliable reconstruction in sparse-view settings, incorporating data-driven priors has become essential, as analytical and optimization-based methods alone cannot recover fine structural details from limited measurements. To this end, neural-network-based approaches have been explored to learn such priors directly from data~\cite{lin2023learning, lin2024learning, kyung2023perspective, shen2019patient, liu2020tomogan, ying2019x2ct, adler2018learned, lin2024c}. With the growing availability of large-scale datasets and increased computational capacity, recent feed-forward models~\cite{zhang2025x, liu2025x} have shown strong potential for sparse-view CT reconstruction. However, their results still suffer from artifacts and a loss of fine details, suggesting that further advances are needed for accurate and reliable volumetric recovery.

Although these methods employ 3D representations, such as triplanes or voxel-aligned Gaussian primitives, they do not explicitly maintain a persistent 3D latent representation that serves as the core reconstruction state. Without a dedicated 3D reconstruction state, ensuring globally consistent volumetric structure across the entire volume becomes more difficult under sparse-view constraints.
Sparse-view cone-beam CT (CBCT) reconstruction, in contrast, fundamentally seeks to recover a geometrically and physically coherent 3D attenuation volume from limited projection measurements. This objective calls for architectural designs that operate directly on an explicit volumetric representation and support coherent integration of multi-view information within a unified 3D space.

In this work, we propose a novel feed-forward framework that integrates large-scale data-driven prior learning with classical iterative reconstruction principles, achieving fast and accurate CBCT reconstruction and novel-view X-ray synthesis from sparse multi-view projections. We introduce an Iterative Latent Volume (\textit{ILV}), which serves as an explicit 3D latent representation that is iteratively refined through the interaction between the input X-ray projections and the learned data-driven prior. By combining the complementary strengths of both paradigms, the proposed framework captures fine structural details through iterative refinement while leveraging learned priors that provide both local structural cues and global anatomical context, enabling the \textit{ILV} to reconstruct coherent 3D volumes even under sparse-view conditions. As a result, the model effectively integrates multi-view information into the 3D latent representation, yielding high-fidelity reconstructions and consistent novel-view synthesis.

To embody the proposed concept, we develop a series of architectural modules. The first is the X-ray feature volume, which is geometrically aligned with the latent volume. For each input X-ray view, the model extracts both high-level semantic and low-level structural features and back-projects them into a shared 3D space, allowing direct integration of view-specific X-ray information into the latent volume.
Second, we incorporate group cross-attention to efficiently model the interaction between the X-ray feature volume and the latent volume. Since the computational cost of full cross-attention grows cubically with the spatial resolution, we divide the 3D space into local groups and compute attention within each group, substantially reducing complexity while preserving effective feature exchange across volumes.
Third, because the latent volume contains a large number of tokens (e.g., $32^3$), applying standard self-attention would be computationally prohibitive. To address this, we employ an efficient attention mechanism that reduces computational overhead while maintaining long-range dependencies for effective latent refinement. Fourth, we introduce a view-wise X-ray feature aggregation module that provides complementary information across multiple views to enhance the global consistency of the reconstructed volume. Finally, the refined latent volume is decoded into Gaussian parameters, from which both novel-view X-ray projections and 3D CT volumes can be synthesized.

The proposed \textit{ILV} is evaluated on approximately 14,000 CT volumes and consistently outperforms both conventional and learning-based methods. 
It improves PSNR by 6\,dB over DIF-Gaussian under 10 views and surpasses R$^{2}$-Gaussian by over 2.6\,dB under 24 views, while reconstructing each case within one second. 
We believe our findings open up new possibilities toward fast, accurate, and near real-time sparse-view CT reconstruction in future clinical and research applications.

\section{Related work}
\label{sec:related_work}

\noindent{\textbf{Feed-forward 3D reconstruction.}}
Feed-forward 3D reconstruction aims to predict scene geometry, camera poses, and novel views from a few input images in a single forward pass, without any scene-specific optimization. Recent approaches can be categorized by their scene representation into pointmap-based~\cite{wang2024dust3r, leroy2024grounding, yang2025fast3r, wang2025continuous, wang2025vggt}, 3D Gaussian Splatting-based~\cite{szymanowicz2024splatter, xu2024grm, zhang2024gs, ziwen2025long, charatan2024pixelsplat, xu2025depthsplat, kang2025selfsplat, kang2025ilrm, ye2024no}, 3D-free methods~\cite{jin2024lvsm, jiang2025rayzer} and others. Among these, feed-forward 3DGS methods leverage priors learned from large-scale datasets to infer Gaussian parameters from only a few views, enabling efficient novel-view synthesis. Some methods~\cite{zhang2024gaussiancube, jiang2025anysplat, chen2024lara} adopt a volumetric representation, predicting 3D Gaussians within an explicit 3D grid. Building on the feed-forward 3D reconstruction paradigm, our method introduces a CT-tailored architecture that models the volumetric density field and embeds iterative reconstruction principles into a feed-forward framework.

\noindent{\textbf{Sparse-view CT reconstruction.}}
Sparse-view CT reconstruction aims to recover a 3D CT volume from a limited number of 2D projections.
Early studies primarily relied on analytical and iterative reconstruction algorithms~\cite{feldkamp1984practical, yu2006region, sauer2002local, andersen1984simultaneous, pan2006variable, sidky2008image}.
Analytical approaches~\cite{feldkamp1984practical, yu2006region} are computationally efficient but degrade rapidly with limited projections, producing streak artifacts. Iterative methods~\cite{andersen1984simultaneous, pan2006variable, sidky2008image, sauer2002local} better preserve image quality with fewer views but are computationally expensive and unstable under extremely sparse-view conditions.
Recent advances in deep learning have accelerated its application to sparse-view CT reconstruction, mainly through optimization-based and feed-forward approaches.
Optimization-based methods~\cite{cai2024radiative, cai2024structure, zang2021intratomo, ruckert2022neat, zhaneural, zha2024r, shen2022nerp} using NeRF~\cite{mildenhall2021nerf} and 3DGS~\cite{kerbl3Dgaussians} can achieve high-quality reconstructions from limited projections, but require per-scene optimization and long runtimes.
Feed-forward methods~\cite{lin2023learning, lin2024learning, kyung2023perspective, shen2019patient, liu2020tomogan, ying2019x2ct, adler2018learned, lin2024c} enable fast inference but often generalize poorly due to limited model capacity and insufficient data diversity.
While diffusion-based variants~\cite{lee2023improving, chung2023decomposed, chung2023solving} have been introduced to improve reconstruction quality, they further increase inference time due to multi-step sampling.
Most recently, large-scale feed-forward models~\cite{liu2025x, zhang2025x} have shown strong potential for sparse-view CT reconstruction, yet their results still suffer from artifacts and a loss of fine details. Therefore, further advances are needed for accurate and reliable volumetric recovery.
\vspace{-0.2cm}

\section{Method}
\label{sec:method}

\begin{figure*}[t]  
    \centering
    \includegraphics[width=1\textwidth]{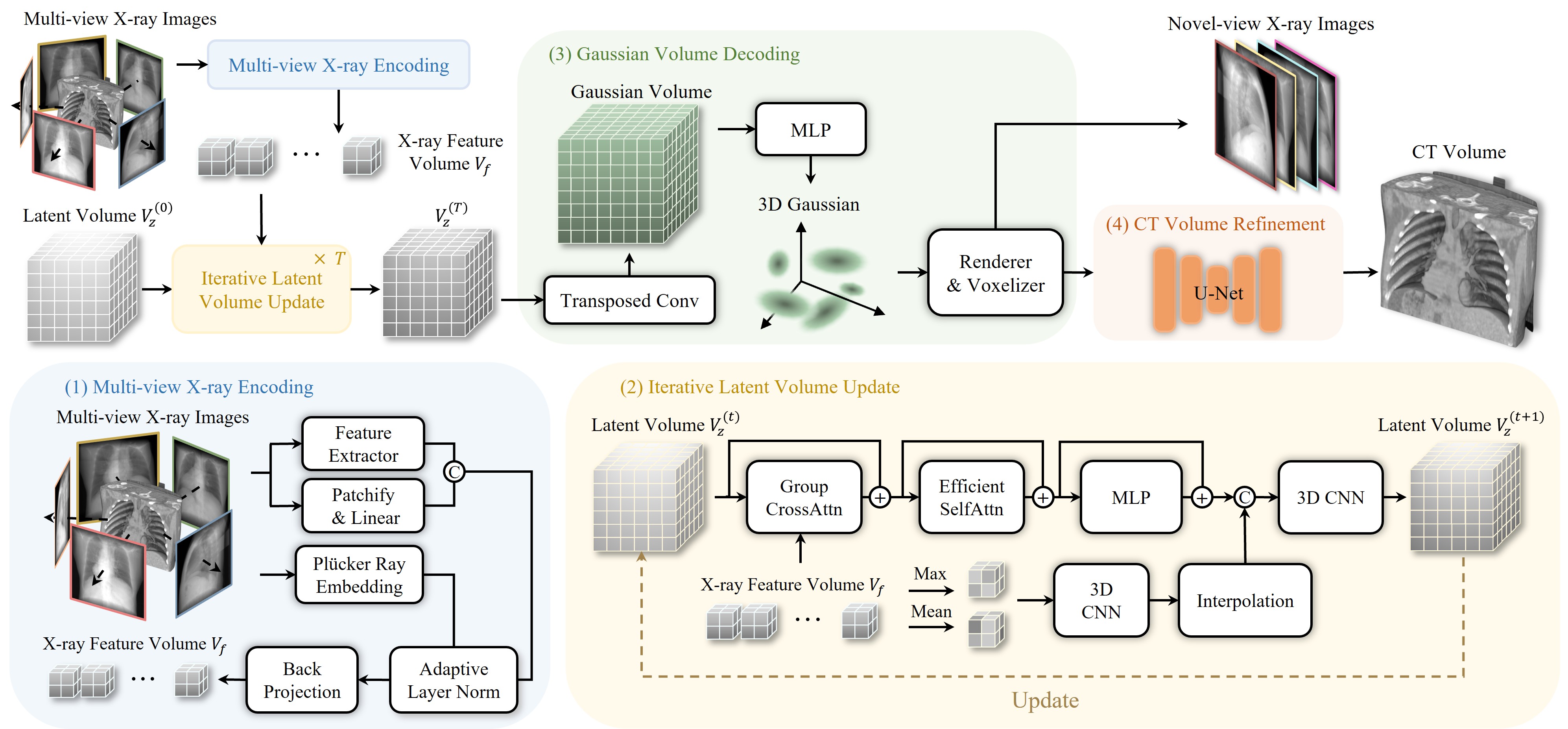} 
    \caption{Overview of the proposed \textit{ILV}.
Given multi-view X-ray images, \textit{ILV} reconstructs a 3D CT volume or synthesizes novel-view projections. The overall network consists of four stages: (1) Multi-view X-ray image encoding, (2) Latent volume update, (3) Gaussian volume decoding, and (4) CT volume refinement.}
    \label{fig:main}
\end{figure*}

\subsection{Overview}
We propose \textit{ILV}, a model that takes multi-view X-ray images as input and can both reconstruct a 3D CT volume or synthesize X-ray projections from novel viewpoints.
The model first extracts hybrid features from the X-ray images and lifts them into 3D space to construct an X-ray feature volume. 
The X-ray feature volume is repeatedly injected into a learnable latent volume, and the latent volume itself is refined internally at each iteration.
Through the iterative injection-refinement process within the 3D latent volume, the model progressively learns structural priors over the CT volume. 
As shown in~\cref{fig:main}, the overall network consists of four stages: (1) Multi-view X-ray image encoding, (2) Latent volume update, (3) Gaussian volume decoding, and (4) CT volume refinement.

\subsection{Multi-view X-ray Encoding}
Given $N$ input X-ray projection images, we extract a high-level feature map $\{ f_\text{high}^{(i)} \}_{i=1}^N$ using a ViT backbone~\cite{dosovitskiy2020image} and a low-level feature map $\{ f_\text{low}^{(i)} \}_{i=1}^N$ using a patchify-linear projection. These two feature maps are then concatenated to form a hybrid feature map $f_\text{hybrid}^{(i)} = \operatorname{concat}\big(f_\text{high}^{(i)}, f_\text{low}^{(i)}\big)$.
By jointly exploiting high-level semantic and low-level structural cues, the model can capture both global context and local details. 
For each view $i$, we compute the Pl\"ucker ray embedding $R^{(i)}$ of the X-rays, and we adopt adaptive layer normalization to inject conditional information~\cite{peebles2023scalable}.
The resulting multi-view hybrid features are then back-projected into 3D space to construct per-view X-ray feature volumes $V_{f}^{(i)} \in \mathbb{R}^{C_f \times L_f \times L_f \times L_f}$.
Specifically, the BackProj operation projects each voxel’s 3D position onto the 2D image plane, samples the corresponding 2D feature, and assigns it back to the voxel to form a 3D feature volume.
All per-view feature volumes are subsequently stacked along the view dimension to obtain the global X-ray feature volume $V_{f} \in \mathbb{R}^{N \times C_f \times L_f \times L_f \times L_f}$:
\begin{align}
    V_{f}^{(i)} &= \operatorname{BackProj}\big(\operatorname{AdaLN}(f^{(i)}_{\text{hybrid}}, R^{(i)})\big), \\
    V_{f} &= \operatorname{Stack}\big(\{V_{f}^{(i)}\}_{i=1}^N\big),
\end{align}
where $\operatorname{AdaLN}$ denotes the adaptive layer normalization. The lifted 3D features establish an aligned representation that bridges the 2D X-ray observations and the latent volumetric space. The 3D X-ray feature volume $V_{f}$ is iteratively injected into the latent volume, enabling the network to integrate multi-view geometric information and effectively exploit the X-ray observations.

\subsection{Latent Volume Update}

In this stage, we introduce a learnable 3D latent volume, denoted as 
$V_z^{(0)} \in \mathbb{R}^{C_z \times L_z \times L_z \times L_z}$.
This volume acts as a learnable 3D feature representation that stores and
progressively refines structural information across stages.
The latent volume  is iteratively updated at each stage $t$ by injecting multi-view information from the X-ray feature volume $V_{f}$ and refining its internal structure, allowing it to gradually acquire a CT-like structural representation.

Applying full attention to 3D volumes is computationally expensive due to the cubic growth of tokens with spatial resolution.
To mitigate this, we first perform group cross-attention~\cite{chen2024lara} on the current latent volume $V_z^{(t)}$
conditioned on the X-ray feature volume $V_{f}$.
Specifically, the 3D space is partitioned into multiple 3D spatial groups, and
cross-attention is applied in parallel for each group, enabling efficient
multi-view information injection. This strategy not only reduces computational overhead but also respects geometric locality and regional correspondence, thereby efficiently integrating geometry-aware multi-view X-ray information into the latent volume.
\begin{equation}
    \hat{V}_z^{(t)} 
    = \operatorname{GroupCrossAttn}\big(\operatorname{LN}(V_z^{(t)}),\, V_{f}\big) + V_z^{(t)}. 
    \label{eq:latent_cross}
\end{equation}
We then refine the internal structure of the latent volume using an Efficient Attention mechanism~\cite{xie2021segformer, perera2024segformer3d} followed by an MLP. Both operations are applied within the latent volume to model spatial dependencies among voxels:
\begin{equation}
    \tilde{V}_z^{(t)} 
    = \hat{V}_z^{(t)} + \operatorname{EffAttn}(\hat{V}_z^{(t)}).
    \label{eq:latent_effattn}
\end{equation}
\begin{equation}
    \bar{V}_z^{(t)} 
    = \tilde{V}_z^{(t)} + \operatorname{MLP}(\tilde{V}_z^{(t)}).
    \label{eq:latent_mlp}
\end{equation}
Unlike full self-attention with a quadratic complexity of $O(N^2)$, where $N = L_z^3$ represents the sequence length of all latent volume tokens, Efficient Attention reduces the key and value sequence length by a factor of $R$, yielding an approximate complexity of $O(N^2 / R)$.
This allows the latent volume to capture global context and maintain structural consistency while significantly reducing computational overhead. As a result, the model can effectively refine volumetric representations without compromising long-range dependencies.

Finally, to further inject multi-view X-ray information while preserving a coherent 3D representation, we incorporate  view-wise mean-max features from the X-ray feature volume $V_{f}$ into the latent volume. Specifically, we aggregate $V_f$ across the view dimension using mean and max pooling, followed by $1 \times 1 \times 1$ 3D convolutions and spatial interpolation to match the latent resolution.
The resulting features are concatenated with $\bar{V}_z^{(t)}$ along the channel dimension and processed by a 3D convolution to produce the next-stage latent volume:
\begin{equation}
    V_z^{(t+1)} 
    = \operatorname{Conv3D}\big(
        [\bar{V}_z^{(t)},\, \text{mean}(V_f),\, \text{max}(V_f)]
    \big).
    \label{eq:latent_update}
\end{equation}
By combining multi-view information through mean and max operations, the model is encouraged to preserve consistent structural cues and to aggregate complementary information from multi-views.
In summary, the Latent Volume Update stage consists of
(1) group cross-attention for multi-view X-ray injection,
(2) efficient self-attention and MLP for internal structural refinement, and
(3) view-wise mean–max aggregation for consolidating a consistent 3D representation.
Through these steps, the latent volume progressively resembles the structural representation of a CT volume.

\subsection{Gaussian Volume Decoding}

In the Gaussian Volume Decoding stage, the refined latent volume $V_z^{(T)}$ is upsampled by a 3D transposed convolution to generate a high-resolution Gaussian feature volume:
\begin{equation}
    G = \operatorname{ConvTranspose3D}(V_z^{(T)}; k, s),
    \label{eq:gaussian_feat_vol}
\end{equation}
where $G$ denotes the decoded 3D Gaussian feature volume, and $k$ and $s$ represent the kernel size and stride that together determine the upsampling ratio of the transposed convolution.
For each voxel $j$ in the 3D grid, the corresponding feature vector $G_j$ is decoded to Gaussian primitive parameters by an MLP-based decoder:
\begin{equation}
    \{o_j, s_j, r_j, d_j\}
    = \operatorname{MLP}(G_j),
    \label{eq:gaussian_decoding}
\end{equation}
where $o_j$ is the voxel position offset, $s_j$ is the scale, $r_j$ represents the rotation and $d_j$ denotes the density.
We adopt the differentiable Gaussian renderer and voxelizer from R\textsuperscript{2}-Gaussian~\cite{zha2024r} to synthesize X-ray projections from novel viewpoints and reconstruct 3D CT volumes from the predicted Gaussian parameters.

\subsection{CT Volume Refinement}
Although the coarse CT volume reconstructed from the predicted Gaussian primitives captures the overall anatomical structure, it may still exhibit blurred tissue boundaries and locally abnormal intensity patterns. To alleviate these artifacts, we introduce a lightweight 3D U-Net module that performs residual refinement of the coarse CT volume.
The module consists of two downsampling stages (with factors of $4$ and $2$) followed by symmetric upsampling stages, forming an encoder-decoder architecture with skip connections.
The refined CT volume $V_{\text{refined}}$ is obtained in a residual form with respect to the coarse volume $V_{\text{coarse}}$:
\begin{equation}
    V_\text{refined} = V_{\text{coarse}} + \operatorname{UNet3D}(V_{\text{coarse}}).
    \label{eq:ct_refine}
\end{equation}
This refinement stage helps correct local intensity inconsistencies and recover sharper structural details, leading to more stable and high-fidelity CT reconstructions with notable performance gains.

\subsection{Loss Function}
We train the model to jointly optimize 3D CT reconstruction and novel-view X-ray rendering by combining three loss terms. The volumetric reconstruction loss encourages the predicted coarse CT volume $V_{\text{coarse}}$ to match the ground-truth CT volume, $\mathcal{L}_{\text{vol}} = \operatorname{MSE}(V_{\text{coarse}}, V_{\text{gt}})$. The rendering loss enforces consistency between the rendered X-ray image $I$ and the ground-truth image $I_{\text{gt}}$:
\begin{equation}
    \mathcal{L}_{\text{img}}
    = \lambda_{\text{L1}} \,\mathcal{L}_{\text{L1}}(I, I_{\text{gt}})
    + \lambda_{\text{SSIM}} \,\mathcal{L}_{\text{SSIM}}(I, I_{\text{gt}}).
\end{equation}
The refinement loss is applied to the refined CT volume $V_{\text{refine}}$ and is activated only after a certain number of training iterations, once the coarse CT reconstruction has become sufficiently stable, $\mathcal{L}_{\text{refined}}=\operatorname{MSE}(V_{\text{refined}}, V_{\text{gt}})$. The total training objective is the sum of the three terms:
\begin{equation}
    \mathcal{L}_{\text{total}}
    = \mathcal{L}_{\text{img}}
    + \lambda_{\text{vol}}\mathcal{L}_{\text{vol}}
    + \lambda_{\text{refined}}\mathcal{L}_{\text{refined}}.
\end{equation}

\section{Experiment}
\label{sec:experiment}

\subsection{Experimental Settings}
\noindent{\textbf{Dataset and metrics.}}
Similar to previous works~\cite{zhang2025x, liu2025x}, we curate a dataset comprising 14,326 CT volumes collected from six publicly available datasets: AbdomenAtlas~\cite{li2024abdomenatlas}, RSNA2023~\cite{hermans2024rsna}, AMOS~\cite{ji2022amos}, MELA~\cite{mela_dataset}, LUNA16~\cite{setio2017validation}, and RibFrac~\cite{jin2020deep, yang2025deep}.
The dataset covers diverse anatomical regions such as the chest, abdomen, and pelvis. 
We resample all CT volumes to a uniform isotropic voxel spacing and resize them to $256^3$ voxels. 
Hounsfield Unit (HU) values are normalized from the range of -1000 $\sim$ 1000 to 0 $\sim$ 1. 
To ensure data reliability, we remove volumes that exhibit abnormal intensity distributions or outlier values.
The preprocessed dataset is divided into three subsets: 
13{,}000 volumes for training, 653 volumes for validation, and 653 volumes for testing. 
Subsequently, synthetic X-ray projection images are generated from each CT volume using the TIGRE toolbox~\cite{biguri2016tigre}. For every CT volume, 50 projections with a resolution of $512^2$ pixels are rendered over the full angular range from $0^{\circ}$ to $360^{\circ}$.

For quantitative evaluations (\cref{tab:tab1}–\cref{tab:tab6}), PSNR is computed in 3D over reconstructed volumes, and SSIM is averaged from 2D scores across axial, coronal, and sagittal slices. For novel-view X-ray synthesis (\cref{tab:tab7}), both metrics are computed in 2D image space.
\medskip

\begin{table}[t]
\caption{Quantitative comparison with traditional and feed-forward methods on CT reconstruction. Runtime is measured with 10 input views.}
\centering
\resizebox{\textwidth}{!}{%
\setlength{\tabcolsep}{10pt}
\begin{tabular}{@{}ccc *{6}{C{1.15cm}}@{}}
\toprule[0.15em]
\multirow{2}{*}{Type} 
& \multirow{2}{*}{Method} 
& \multirow{2}{*}{Time $\downarrow$} 
& \multicolumn{2}{c}{6-View} 
& \multicolumn{2}{c}{8-View} 
& \multicolumn{2}{c}{10-View} \\

\cmidrule(lr){4-5} \cmidrule(lr){6-7} \cmidrule(lr){8-9}

& & 
& PSNR$\uparrow$ & SSIM$\uparrow$ 
& PSNR$\uparrow$ & SSIM$\uparrow$ 
& PSNR$\uparrow$ & SSIM$\uparrow$ \\

\midrule
\multirow{3}{*}{Traditional} 
& FDK~\cite{feldkamp1984practical} & \textbf{0.23s} & 12.79 & 0.122 & 14.58 & 0.145 & 14.75 & 0.166 \\ 
& ASD-POCS~\cite{sidky2008image} & 1m 32s & 22.48 & 0.661 & 23.57 & 0.695 & 24.37 & 0.721 \\ 
& SART~\cite{andersen1984simultaneous} & 2m 48s & 23.21 & 0.689 & 24.26 & 0.712 & 25.06 & 0.733 \\ 

\midrule
\multirow{1}{*}{2D feed-forward} 
& FreeSeed~\cite{ma2023freeseed} & 4.5s & 28.81 & 0.793 & 29.61 & 0.833 & 30.34 & 0.837 \\ 

\midrule
\multirow{4}{*}{3D feed-forward} 
& DIF-Net~\cite{lin2023learning} & 3.0s & 24.18 & 0.720 & 24.59 & 0.734 & 24.65 & 0.745 \\
& DIF-Gaussian~\cite{lin2024learning} & 3.0s & 26.56 & 0.810 & 27.46 & 0.829 & 27.88 & 0.837 \\
& DeepSparse~\cite{lin2026deepsparse} & 12.0s & 30.05 & 0.874 & 31.12 & 0.886 & 31.78 & 0.895 \\
& Ours & 0.59s & \textbf{33.45} & \textbf{0.922} & \textbf{33.25} & \textbf{0.919} & \textbf{33.84} & \textbf{0.924} \\ 

\bottomrule[0.15em]
\end{tabular}}
\label{tab:tab1}
\end{table}

\begin{figure}[t]
\vspace{-0.4cm}
  \centering
  \includegraphics[width=1\textwidth]{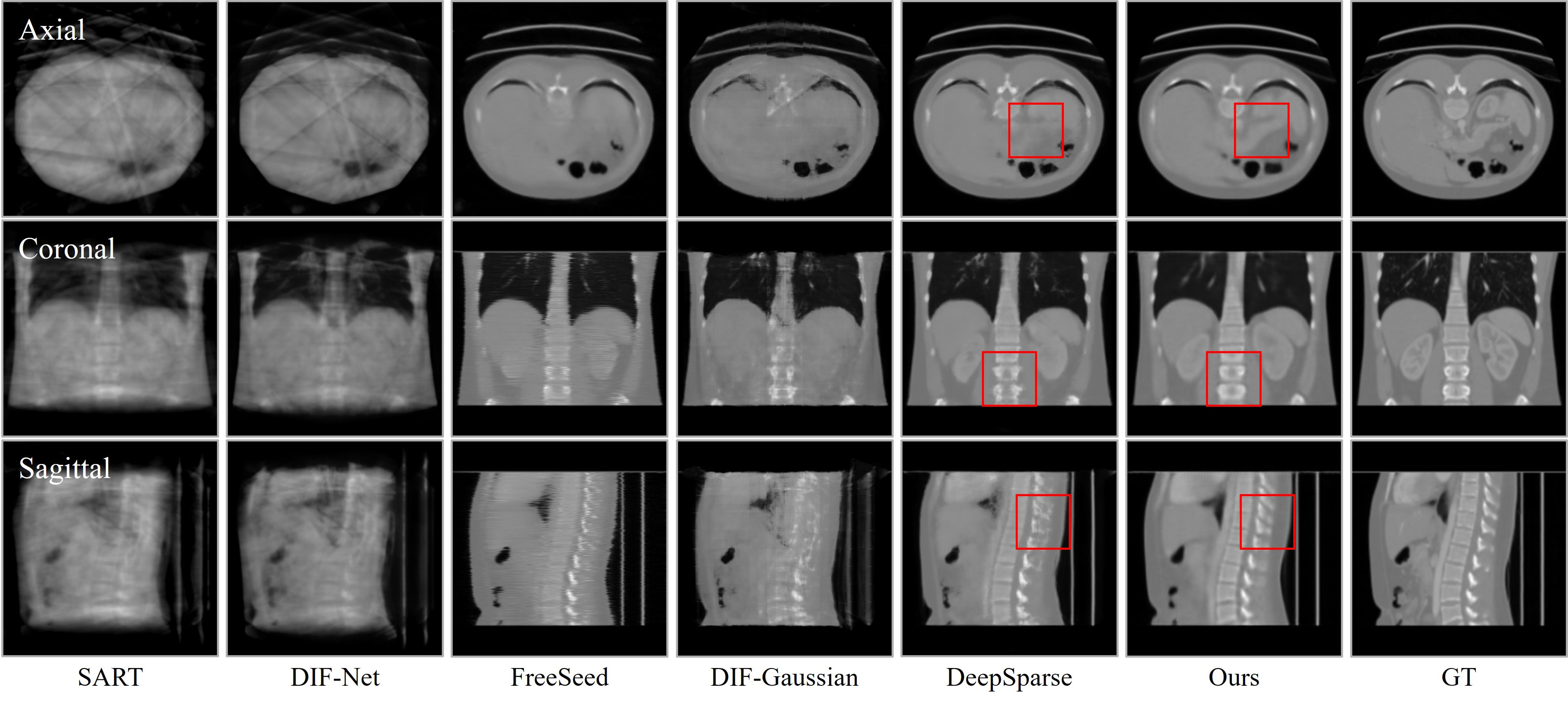}
  \caption{Qualitative comparison with traditional and feed-forward methods on CT reconstruction in the 10-view setting. }
  \label{fig2}
\vspace{-0.1cm}
\end{figure}

\noindent{\textbf{Implementation Details.}}
Our model is implemented using the PyTorch Lightning framework~\cite{falcon2019pytorch}. 
Training is conducted on 8 H100 GPUs with a batch size of 1 per GPU. 
The X-ray feature volume $V_{f}$ has a spatial resolution of $L_f=16$ and $C_f=1536$ channels, 
while the latent volume $V_{z}$ has a resolution of $L_z=32$ and $C_z=256$ channels.
The high-level feature encoder employs a DINO ViT backbone~\cite{caron2021emerging}, group cross-attention uses 16 groups, and efficient self-attention reduces key–value tokens by a factor of $R=8$.
The network consists of 18 layers, followed by a ConvTranspose3D layer with a kernel size of $4$ and a stride of $4$,
producing a Gaussian volume of size $128\times128\times128\times60$.
Each voxel corresponds to a single primitive, and the offset radius is limited to $\frac{1}{64}$ of the bounding box length. 
The total number of trainable parameters is 185M.
We employ the AdamW optimizer~\cite{loshchilov2017decoupled} with momentum coefficients $\beta_1=0.9$ and $\beta_2=0.95$. 
The initial learning rate is set to $1\times10^{-4}$, and the model is trained for 50 epochs.
A cosine warm-up scheduler with 1000 warm-up steps is applied,
and the refinement loss is activated after 5000 training iterations.
\medskip

\begin{table}[t]
\caption{Quantitative comparison with optimization-based methods on CT reconstruction. Runtime is measured with 24 input views.}
\centering
\resizebox{\textwidth}{!}{%
\setlength{\tabcolsep}{12pt}
\begin{tabular}{@{}cc *{6}{C{1.15cm}}@{}}
\toprule[0.1em]
\multirow{2}{*}{Method} 
& \multirow{2}{*}{Time$\downarrow$} 
& \multicolumn{2}{c}{6-View} 
& \multicolumn{2}{c}{10-View} 
& \multicolumn{2}{c}{24-View} \\ 

\cmidrule(lr){3-4} \cmidrule(lr){5-6} \cmidrule(lr){7-8}

& 
& PSNR$\uparrow$ & SSIM$\uparrow$ 
& PSNR$\uparrow$ & SSIM$\uparrow$ 
& PSNR$\uparrow$ & SSIM$\uparrow$ \\ 

\midrule 
IntraTomo~\cite{zang2021intratomo} 
& 9m 26s 
& 24.49 & 0.722 
& 26.30 & 0.772 
& 28.62 & 0.837 \\ 

NAF~\cite{zhaneural} 
& 5m 04s 
& 23.74 & 0.678 
& 26.17 & 0.741 
& 31.34 & 0.876 \\ 

SAX-NeRF~\cite{cai2024structure} 
& 1h 35m 
& 24.58 & 0.754 
& 26.78 & 0.794 
& 33.14 & 0.919 \\ 

R\textsuperscript{2}-Gaussian~\cite{zha2024r} 
& 17m 37s 
& 24.54 & 0.773 
& 27.26 & 0.823 
& 33.29 & 0.931 \\ 

\midrule
Ours 
& \textbf{0.76s} 
& \textbf{33.57} & \textbf{0.923} 
& \textbf{33.95} & \textbf{0.925} 
& \textbf{35.93} & \textbf{0.941} \\ 

\bottomrule[0.1em]
\end{tabular}%
}
\label{tab:tab2}
\end{table}

\begin{figure}[t]
\vspace{-0.3cm}
  \centering
  \includegraphics[width=1\textwidth]{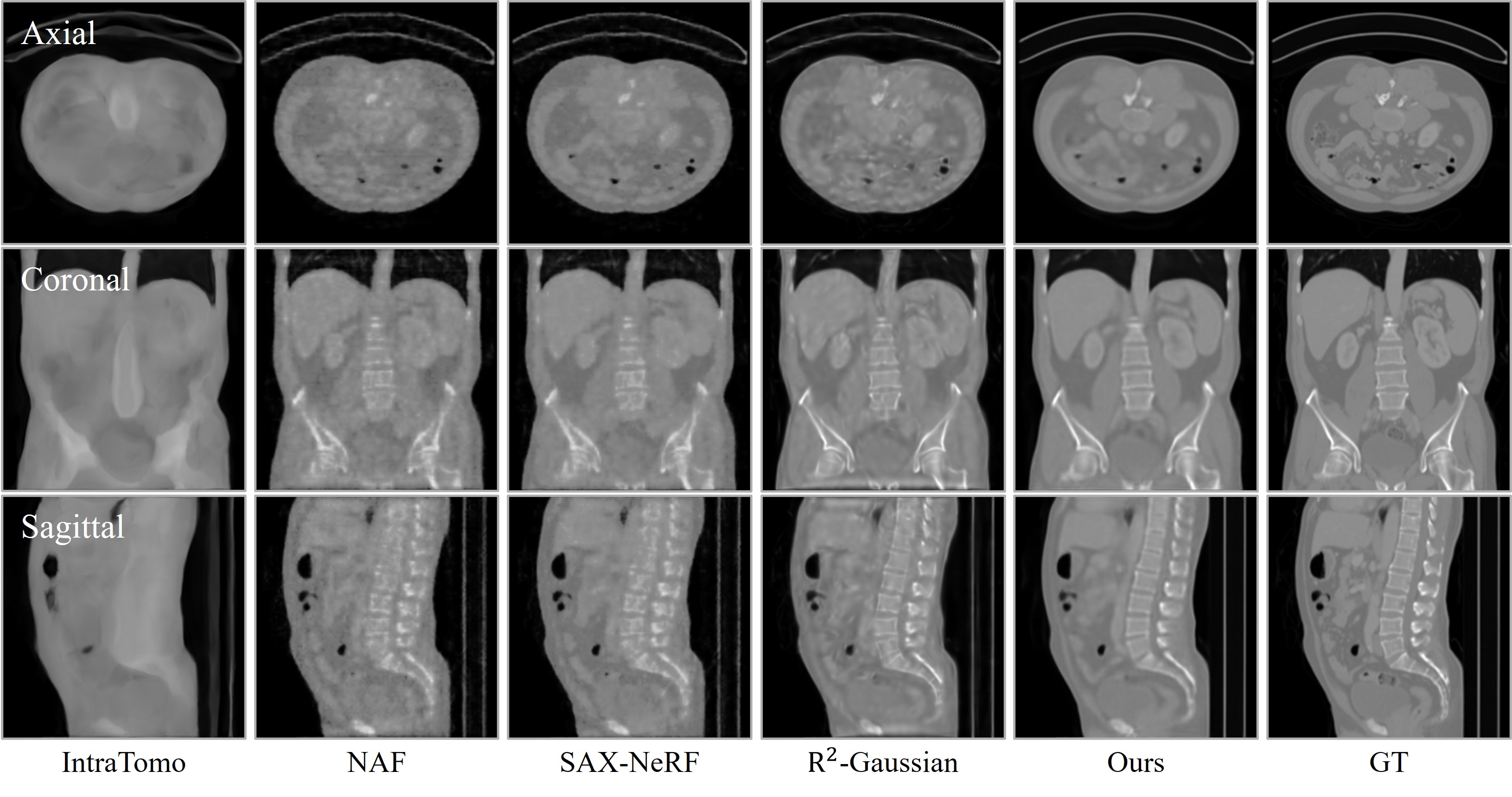} 
  \caption{Qualitative comparison with optimization-based methods on CT reconstruction in the 24-view setting.}
  \label{fig4}
\end{figure}

\subsection{Evaluation}
In this section, we compare the proposed method with existing CT reconstruction approaches, including traditional, feed-forward, and optimization-based methods. 
For a fair comparison, all feed-forward baselines, including FreeSeed~\cite{ma2023freeseed}, DIF-Net~\cite{lin2023learning}, DIF-Gaussian~\cite{lin2024learning}, and DeepSparse~\cite{lin2026deepsparse} are retrained on our dataset\footnote{The original baseline models were trained on relatively small datasets; we retrain them on our large-scale dataset to ensure a fair comparison.}. Runtime is measured on an RTX 4090 GPU and reported as the average inference time over all test samples.

\cref{tab:tab1} summarizes the CT reconstruction performance of our method compared with traditional and feed-forward baselines. We evaluate 653 test volumes and report the average PSNR and SSIM for 6, 8, and 10-view input configurations. 
\cref{tab:tab2} presents comparisons with optimization-based methods on 50 randomly selected test samples under 6, 10, and 24-view settings. Since optimization-based methods typically benefit from denser projections, we additionally include the 24-view configuration.
Note that recent methods~\cite{zhang2025x, liu2025x} are not included, as their official implementations have not been released.

\begin{table*}[t]
\centering
\scriptsize
\setlength{\tabcolsep}{0pt}
\renewcommand{\arraystretch}{0.98}
\begin{minipage}[t]{0.49\textwidth}
\centering
\caption{Cross-dataset evaluation on the 6-view setting.}
\label{tab:tab3}
\vspace{-0.2cm}
\begin{tabular}{>{\raggedright\arraybackslash}p{2.3cm}cccc}
\toprule[0.1em]
\multirow{2}{*}{Method} & \multicolumn{2}{c}{FUMPE~\cite{masoudi2018new}} & \multicolumn{2}{c}{PENGWIN~\cite{liu2023pelvic}} \\
\cmidrule{2-3} \cmidrule{4-5} 
& PSNR$\uparrow$ & SSIM$\uparrow$ & PSNR$\uparrow$ & SSIM$\uparrow$ \\
\midrule  
FreeSeed~\cite{ma2023freeseed}      & 25.95 & 0.776 & 27.79 & 0.792 \\
DIF-Net~\cite{lin2023learning}       & 21.18 & 0.690  & 24.56 & 0.750\\
DIF-Gaussian~\cite{lin2024learning}  & 23.32 & 0.783 & 26.48 & 0.824 \\  
DeepSparse~\cite{lin2026deepsparse}  & 26.85 & 0.839 & 29.17 & 0.875 \\   
NAF~\cite{zhaneural}   & 22.73 & 0.689 & 25.65 & 0.744  \\   
SAX-NeRF~\cite{cai2024structure}      & 22.22 & 0.708 & 25.93   & 0.795\\    
R$^2$-Gaussian~\cite{zha2024r}& 22.45 & 0.723 & 25.89 & 0.805 \\
Ours          & \textbf{30.66} & \textbf{0.909} & \textbf{33.01} & \textbf{0.934} \\
\bottomrule[0.1em]
\end{tabular}
\end{minipage}
\hfill
\begin{minipage}[t]{0.49\textwidth}
\centering
\caption{Cross-dataset evaluation on the 10-view setting.}
\label{tab:tab4}
\vspace{-0.2cm}
\begin{tabular}{>{\raggedright\arraybackslash}p{2.3cm}cccc}
\toprule[0.1em]
\multirow{2}{*}{Method} & \multicolumn{2}{c}{FUMPE~\cite{masoudi2018new}} & \multicolumn{2}{c}{PENGWIN~\cite{liu2023pelvic}} \\
\cmidrule{2-3} \cmidrule{4-5} 
& PSNR$\uparrow$ & SSIM$\uparrow$ & PSNR$\uparrow$ & SSIM$\uparrow$ \\
\midrule  
FreeSeed~\cite{ma2023freeseed}      & 27.12 & 0.801 & 29.98 & 0.841 \\
DIF-Net~\cite{lin2023learning}       & 21.24 & 0.692  & 24.65 & 0.751\\
DIF-Gaussian~\cite{lin2024learning}  & 24.38 & 0.802 & 27.32 & 0.838 \\    
DeepSparse~\cite{lin2026deepsparse}  & 28.06 & 0.859 & 30.70 & 0.893 \\
NAF~\cite{zhaneural}      & 25.19 & 0.794 & 27.85 & 0.796 \\   
SAX-NeRF~\cite{cai2024structure}      & 24.73 & 0.759 & 28.67 & 0.839  \\    
R$^2$-Gaussian~\cite{zha2024r}& 25.11 & 0.785 & 28.36 & 0.853 \\
Ours          & \textbf{31.01} & \textbf{0.911} & \textbf{33.38} & \textbf{0.936} \\
\bottomrule[0.1em]
\end{tabular}
\end{minipage}
\end{table*}

\begin{table*}[t]
\vspace{-0.2cm}
\centering
\scriptsize
\setlength{\tabcolsep}{0pt}
\renewcommand{\arraystretch}{0.98}
\begin{minipage}[t]{0.49\textwidth}
\centering
\caption{Cross-dataset evaluation with feed-forward methods on the 8-view setting.}
\label{tab:tab5}
\vspace{-0.2cm}
\begin{tabular}{>{\raggedright\arraybackslash}p{2.3cm}cccc}
\toprule[0.1em]
\multirow{2}{*}{Method} & \multicolumn{2}{c}{FUMPE~\cite{masoudi2018new}} & \multicolumn{2}{c}{PENGWIN~\cite{liu2023pelvic}} \\
\cmidrule{2-3} \cmidrule{4-5} 
& PSNR$\uparrow$ & SSIM$\uparrow$ & PSNR$\uparrow$ & SSIM$\uparrow$ \\
\midrule  
FreeSeed~\cite{ma2023freeseed}      &26.38 & 0.786 & 28.90 & 0.826 \\
DIF-Gaussian~\cite{lin2024learning}  & 24.03 & 0.795 & 27.00 & 0.837 \\    
DeepSparse~\cite{lin2026deepsparse}  & 27.75 & 0.853 & 30.09 & 0.885 \\
Ours & \textbf{30.49} & \textbf{0.906} & \textbf{32.67} & \textbf{0.932} \\
\bottomrule[0.1em]
\end{tabular}
\end{minipage}
\hfill
\begin{minipage}[t]{0.49\textwidth}
\centering
\caption{Cross-dataset evaluation with optimization-based methods on the 24-view setting.}
\label{tab:tab6}
\vspace{-0.2cm}
\begin{tabular}{>{\raggedright\arraybackslash}p{2.3cm}cccc}
\toprule[0.1em]
\multirow{2}{*}{Method} & \multicolumn{2}{c}{FUMPE~\cite{masoudi2018new}} & \multicolumn{2}{c}{PENGWIN~\cite{liu2023pelvic}} \\
\cmidrule{2-3} \cmidrule{4-5} 
& PSNR$\uparrow$ & SSIM$\uparrow$ & PSNR$\uparrow$ & SSIM$\uparrow$ \\
\midrule  
    NAF~\cite{zhaneural}      & 30.57 & 0.865 & 33.75 & 0.913\\   
    SAX-NeRF~\cite{cai2024structure}      & 31.86 & 0.897 & 35.39   & 0.942 \\    
    R$^2$-Gaussian~\cite{zha2024r}& 31.71 & 0.907 & 34.70 & 0.944 \\
    Ours          & \textbf{33.58} & \textbf{0.935} & \textbf{36.08} & \textbf{0.954} \\[-0.7pt]
\bottomrule[0.1em]
\end{tabular}
\end{minipage}
\vspace{-0.1cm}
\end{table*}

\medskip
\noindent{\textbf{CT Reconstruction.}}
Our model achieves state-of-the-art performance across all view configurations compared with both traditional and feed-forward methods. 
As shown in~\cref{tab:tab1}, it outperforms the 3D feed-forward baseline DeepSparse\cite{lin2026deepsparse} by approximately 2--3\,dB in PSNR. 
Although it is slower than traditional methods such as FDK~\cite{feldkamp1984practical}, the overall inference time remains under one second per case, demonstrating near real-time reconstruction capability. 
Notably, the 6-view configuration yields slightly higher PSNR than the 8-view setting because a higher spatial resolution ($L_f=32$) is used for the X-ray feature volume.

\cref{fig2} shows qualitative results in the 10-view setting for axial, coronal, and sagittal slices, 
comparing our method against both traditional and feed-forward reconstruction approaches. 
FreeSeed~\cite{ma2023freeseed}, which reconstructs 2D slices and then stacks them, exhibits noticeable slice-wise inconsistency in the coronal and sagittal slices, while DeepSparse~\cite{lin2026deepsparse} recovers the overall shape but misses fine details such as bony structures. In comparison, our method better preserves structural details and yields more consistent reconstructions across all three views.

Optimization-based methods are traditionally recognized for their high reconstruction fidelity, albeit at the cost of substantial runtime. In contrast, our method leverages large-scale pretraining to effectively break this conventional trade-off between accuracy and efficiency. 
As shown in \cref{tab:tab2}, our approach consistently achieves the best performance for all view settings. 
In particular, under the 24-view configuration, it surpasses R\textsuperscript{2}-Gaussian~\cite{zha2024r} by about 2.6\,dB in PSNR. 
Moreover, it delivers substantially higher reconstruction quality while being approximately $400\times$ faster than the fastest optimization-based baseline, NAF~\cite{zhaneural}.

\cref{fig4} shows qualitative results in the 24-view setting, comparing our method against optimization-based baselines. 
IntraTomo~\cite{zang2021intratomo} produces overly smoothed reconstructions with blurred organ boundaries. 
NAF~\cite{zhaneural} and SAX-NeRF~\cite{cai2024structure} yield noisy reconstructions and fail to recover fine anatomical details. 
R\textsuperscript{2}-Gaussian \cite{zha2024r} preserves structures more faithfully but exhibits characteristic spiky Gaussian artifacts. 
In contrast, our method produces images that are visually closest to the ground truth, with fewer artifacts across all three views. Additional qualitative results are provided in the supplementary material.
\medskip

\noindent{\textbf{Cross-dataset Generalization.}}
To evaluate cross-dataset generalization, we test our model and several feed-forward and optimization-based baselines on two external datasets, FUMPE~\cite{masoudi2018new} and PENGWIN~\cite{liu2023pelvic}, without any retraining or fine-tuning. 
Results under different view configurations are summarized in \cref{tab:tab3,tab:tab4,tab:tab5,tab:tab6}. 
Our method consistently achieves the best performance across all settings. 
For instance, under the 6-view configuration on FUMPE, it improves PSNR by over 3.8\,dB compared to DeepSparse, and under the 24-view setting, it surpasses R$^2$-Gaussian by up to 1.9\,dB on FUMPE and 1.4\,dB on PENGWIN. 
These results indicate that the proposed latent volume refinement learns structural priors that generalize effectively to unseen domains.

\begin{table}[t]
\centering
\scriptsize

\begin{minipage}[t]{0.42\linewidth}
\vspace{0pt}
\centering
\caption{Quantitative comparison on novel-view X-ray synthesis.}
\label{tab:tab7}
\setlength{\tabcolsep}{6pt}
\resizebox{\linewidth}{!}{%
\begin{tabular}{ccc}
\toprule[0.1em]
Method & PSNR$\uparrow$ & SSIM$\uparrow$ \\
\midrule
IntraTomo~\cite{zang2021intratomo} & 32.03 & 0.980\\
NAF~\cite{zhaneural} & 29.75 & 0.865 \\
SAX-NeRF~\cite{cai2024structure} & 34.27 & 0.990 \\
R\textsuperscript{2}-Gaussian~\cite{zha2024r} & 35.10 & 0.972 \\
Ours & \textbf{43.28} & \textbf{0.992} \\
\bottomrule[0.1em]
\end{tabular}%
}
\end{minipage}
\hfill
\begin{minipage}[t]{0.55\linewidth}
\renewcommand{\arraystretch}{1.025}
\vspace{0pt}
\centering
\caption{Denser-view comparison with R\textsuperscript{2}-Gaussian on CT reconstruction.}
\label{tab:denser_views}
\setlength{\tabcolsep}{4pt}
\resizebox{\linewidth}{!}{%
\begin{tabular}{lcccc}
\toprule[0.1em]
Method & \#Views & Time$\downarrow$ & PSNR$\uparrow$ & SSIM$\uparrow$ \\ 
\midrule
\multirow{3}{*}{R\textsuperscript{2}-Gaussian}
& 24 & 18m & 33.29 & 0.931 \\
& 32 & 19m & 34.78 & 0.952 \\
& 40 & 19m & 35.76 & \textbf{0.963} \\
\midrule
Ours 
& 24 & \textbf{0.76s} & \textbf{35.93} & 0.941 \\
\bottomrule[0.1em]
\end{tabular}%
}
\end{minipage}

\end{table}

\begin{figure}[t]
  \vspace{-0.2cm}
  \centering
  \includegraphics[width=0.65\linewidth]{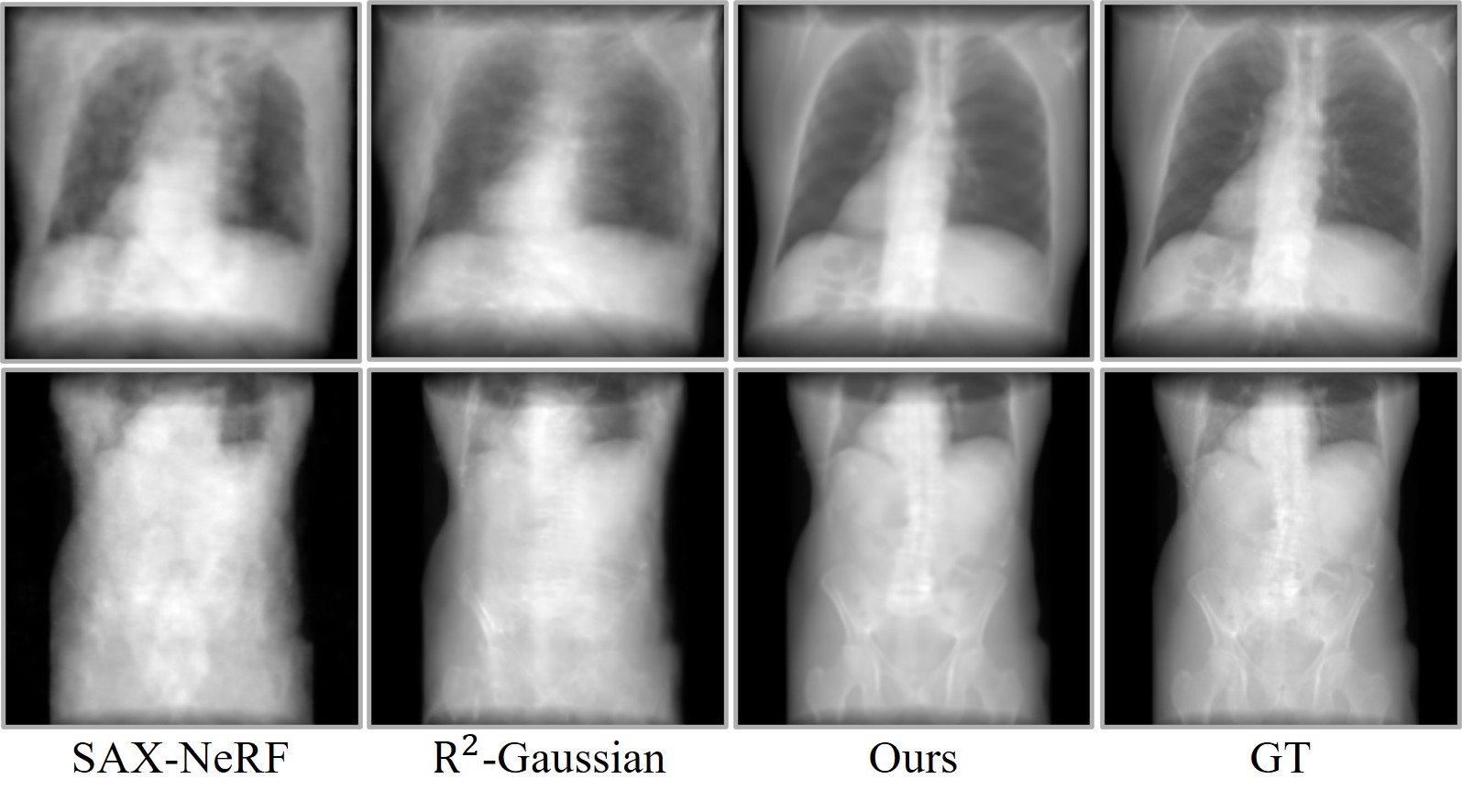}
  \vspace{-0.1cm}
  \caption{Qualitative comparison on novel view X-ray synthesis.}
  \label{fig5}
\end{figure}

\medskip
\noindent{\textbf{Novel View X-ray Synthesis.}}
We further evaluate our method on novel-view X-ray synthesis under a 6-view setting, comparing it with optimization-based methods as summarized in \cref{tab:tab7}. Our model achieves an improvement of about 8\,dB in PSNR over R\textsuperscript{2}-Gaussian~\cite{zha2024r}. As shown in \cref{fig5}, SAX-NeRF~\cite{cai2024structure} and R\textsuperscript{2}-Gaussian tend to produce overly smoothed images with blurred structural boundaries. 
In contrast, \textit{ILV} preserves structural fidelity and recovers clear object boundaries. 
These results demonstrate that our approach not only reconstructs high-quality CT volumes but also performs well in rendering-based applications.

\medskip
\noindent\textbf{Denser-view Comparison.}
Optimization-based methods generally benefit from denser input projections, 
whereas our method aims to achieve high-quality reconstruction with fewer views and sub-second inference.
To examine this trade-off, we additionally evaluate R\textsuperscript{2}-Gaussian~\cite{zha2024r} under 32- and 40-view settings.
As shown in~\cref{tab:denser_views}, increasing the number of projections substantially improves R\textsuperscript{2}-Gaussian.
Nevertheless, our 24-view model achieves slightly higher PSNR than 40-view R\textsuperscript{2}-Gaussian, while maintaining competitive SSIM, requiring fewer projections, and running orders of magnitude faster.
This result shows that our method provides high reconstruction quality with substantially faster inference.

\begin{table}[t]
\centering
\vspace{-0.1cm}
\scriptsize
\caption{Ablation study of \textit{ILV} components.}
\vspace{-0.1cm}
\setlength{\tabcolsep}{12pt}
\begin{tabular}{c cc|cc}
\toprule[0.1em]
\multirow{2}{*}{Method} 
& \multicolumn{2}{c|}{CT Recon.}
& \multicolumn{2}{c}{Novel-view X-ray} \\
\cmidrule(lr){2-3} \cmidrule(lr){4-5}
& PSNR$\uparrow$ & SSIM$\uparrow$
& PSNR$\uparrow$ & SSIM$\uparrow$ \\
\midrule
(a) w/o latent volume 
& 30.07 & 0.875 
& 39.73 & 0.985 \\

(b) Single Inject  
& 31.57 & 0.902
& 41.60 & 0.989 \\

(c) w/o Linear 
& 31.88 & 0.904
& 41.63 & 0.989 \\

(d) w/o UNet3D 
& 31.72 & 0.903
& 41.83 & 0.989 \\

(e) full 
& \textbf{32.08} & \textbf{0.906}
& \textbf{41.83} & \textbf{0.989} \\
\bottomrule[0.1em]
\end{tabular}
\label{tab:tab9}
\end{table}

\begin{table}[t]
\centering
\footnotesize
\caption{Effect of model capacity under the 10-view setting.}
\vspace{-0.1cm}
\setlength{\tabcolsep}{6pt}
\begin{tabular}{cccccc}
\toprule[0.1em]
Method & PSNR$\uparrow$ & SSIM$\uparrow$ & Params & Time \\
\midrule
Ours & 33.84 & 0.924 & 185M & 0.59s \\
Ours-small & 33.52 & 0.921 & 88M & 0.51s \\
\bottomrule[0.1em]
\end{tabular}
\label{tab:model_size}
\end{table}

\subsection{Ablation Studies}
\label{sec:ab}
All ablation models are trained for 30 epochs with 6 input views and $L_f=16$, under the same settings as the main experiment.
As shown in~\cref{tab:tab9}, removing the persistent learnable 3D latent volume (a) causes the largest degradation, reducing CT reconstruction performance by 2.01\,dB in PSNR and 0.031 in SSIM compared to the full model.
This variant uses the same X-ray encoder, Gaussian decoder, and CT refinement module as \textit{ILV}, but directly fuses multi-view X-ray features through repeated self-attention and decodes the fused representation without maintaining a persistent 3D reconstruction state.
The clear performance drop indicates that the main gain of \textit{ILV} comes from preserving a learnable 3D latent state and iteratively injecting multi-view X-ray features into it, rather than relying only on fused 2D X-ray features.
Injecting the X-ray feature volume only at the first stage (b) reduces CT PSNR by 0.51\,dB, indicating that iterative multi-view integration is essential.
Removing low-level features in the multi-view encoding process (c), leaving only the high-level features from the DINO backbone, results in a 0.20\,dB drop, demonstrating that combining low- and high-level features yields richer and more informative representations.
Excluding the 3D U-Net (d) leads to a 0.36\,dB decrease and produces unstable voxel intensities, confirming that the refinement module effectively stabilizes and enhances the coarse CT volumes.
The full model (e) achieves the best overall CT reconstruction performance.

\subsection{Effect of Model Capacity}
To assess the effect of model capacity within our framework, we evaluate a smaller variant of \textit{ILV} under the 10-view setting.
\textit{ILV}-Small adopts a DINO-S backbone with a 12-layer iterative update module, reducing the number of parameters from 185M to 88M.
As shown in~\cref{tab:model_size}, this reduction leads to only a 0.32\,dB PSNR drop while maintaining sub-second inference.
These results suggest that the proposed iterative latent volume design remains effective even with reduced model capacity.

\begin{table}[t]
\centering
\scriptsize
\caption{Quantitative results of perceptual fine-tuning under the 24-view setting.}
\vspace{-0.3cm}
\setlength{\tabcolsep}{14pt}
\begin{tabular}{cccc}
\toprule[0.1em]
Method & PSNR$\uparrow$ & SSIM$\uparrow$ &LPIPS$\downarrow$\\
\midrule
w/o LPIPS & 35.81 & 0.941 & 0.120\\
w/ LPIPS & 35.56 & 0.936 & 0.062  \\
\bottomrule[0.1em]
\end{tabular}
\label{tab:tab10}
\end{table}

\begin{figure}[t]
  \centering
  \vspace{-0.3cm}
  \includegraphics[width=0.65\linewidth]{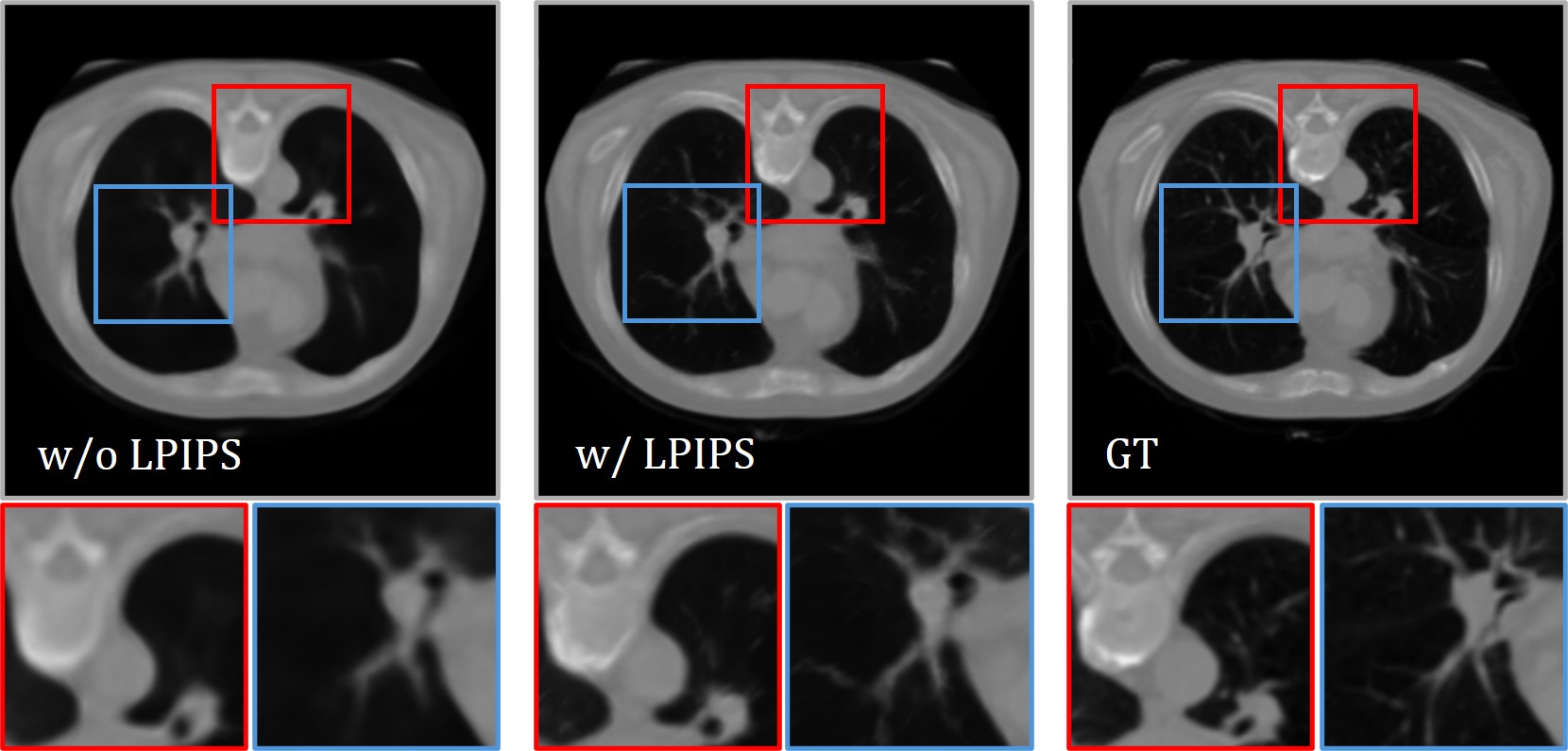}
  \caption{Qualitative results of perceptual fine-tuning.}
  \vspace{-0.4cm}
  \label{fig6}
\end{figure}

\subsection{Fine-tuning for Improving Visual Quality}
\label{sec:finetuning}
Existing methods relying on MSE or MAE losses often produce overly smooth and blurry CT reconstructions.
While \textit{ILV} significantly improves reconstruction quality, it can still exhibit slightly blurred textures in some cases, as shown in~\cref{fig2,fig4}.
To mitigate this limitation, we fine-tune the 24-view model for 10 epochs by incorporating an LPIPS loss computed from randomly sampled 2D slices across all three anatomical axes. This fine-tuning reduces PSNR by approximately 0.25\,dB but cuts the LPIPS score by half, as reported in~\cref{tab:tab10}, leading to sharper and more perceptually detailed reconstructions (see~\cref{fig6}). Minor artifacts still appear occasionally, suggesting that further investigation is needed to better balance perceptual and pixel-wise losses.

\section{Conclusion}
\label{sec:conclusion}
In this work, we introduced Iterative Latent Volumes (\textit{ILV}), a novel feed-forward framework that unifies large-scale data-driven priors with classical iterative reconstruction principles. By progressively refining an explicit 3D latent representation through multi-view X-ray interactions and efficient attention mechanisms, \textit{ILV} enables high-quality and consistent CT reconstruction from sparse-view inputs in near real-time.
Our results demonstrate that incorporating learned priors into an iterative refinement process is a powerful strategy for sparse-view CT.
Extensive experiments show that \textit{ILV} achieves superior reconstruction quality while maintaining sub-second inference time, demonstrating a favorable balance between accuracy and efficiency. We believe that \textit{ILV} opens new opportunities for fast, accurate, and clinically viable volume imaging in future CT systems.

\clearpage  


%
%
\bibliographystyle{splncs04}
\bibliography{main}

\clearpage
\appendix
\setcounter{page}{1}

\begin{center}
\large
\textbf{ILV: Iterative Latent Volumes for Fast and Accurate\\
Sparse-View CT Reconstruction}

\vspace{0.5em}
Supplementary Material
\end{center}
\vspace{1em}

\begin{table}[!h]
\centering

\begin{minipage}{0.6\linewidth}
\centering
\scriptsize
\setlength{\tabcolsep}{0pt}
\caption{Overview of the six public CT datasets and the number of volumes in each dataset.}
\renewcommand{\arraystretch}{1.2}
\begin{tabular}{ccc}
\toprule[0.1em]
Dataset & Body Parts & \# of Volumes \\
\midrule
AbdomenAtlasv1.0~\cite{li2024abdomenatlas} & Abdomen, Chest, Pelvis & 5,171 \\
RSNA2023~\cite{hermans2024rsna} & Abdomen, Pelvis & 4,711 \\
AMOS~\cite{ji2022amos} & Abdomen & 1,851 \\
MELA~\cite{mela_dataset} & Chest & 1,100 \\
LUNA16~\cite{setio2017validation} & Chest & 833 \\
RibFrac~\cite{jin2020deep, yang2025deep} & Abdomen, Chest & 660 \\
\bottomrule[0.1em]
\end{tabular}
\label{tab:dataset}
\end{minipage}
\hfill
\begin{minipage}{0.35\linewidth}
\centering
\scriptsize
\renewcommand{\arraystretch}{1.4}
\caption{Additional Quantitative comparison.}
\begin{tabular}{ccc}
\toprule[0.1em]
Method & PSNR$\uparrow$ & SSIM$\uparrow$ \\
\midrule
IntraTomo~\cite{zang2021intratomo} & 25.63 & 0.752 \\
NAF~\cite{zhaneural} & 25.35 & 0.721 \\
SAX-NeRF~\cite{cai2024structure} & 26.13 & 0.784 \\
R$^2$-Gaussian~\cite{zha2024r} & 26.08 & 0.805 \\
Ours & \textbf{33.35} & \textbf{0.919} \\
\bottomrule[0.1em]
\end{tabular}
\label{tab:8v}
\end{minipage}

\end{table}

\section{Dataset Details}

We collect approximately 14,000 CT volumes from six public datasets: AbdomenAtlas~\cite{li2024abdomenatlas}, RSNA2023~\cite{hermans2024rsna}, AMOS~\cite{ji2022amos}, MELA~\cite{mela_dataset}, LUNA16~\cite{setio2017validation}, and RibFrac~\cite{jin2020deep, yang2025deep}.
All scans undergo a unified preprocessing pipeline to ensure consistent spatial and intensity representation. Each volume is first resampled to a predefined target spacing. After resampling, we standardize the volume size to a resolution of $256^3$.
Depending on the original dimensions, we apply center cropping, padding, or a combination of both. 
If the resulting shape still does not match the target resolution, an additional resizing step is applied to obtain the exact $256^3$ input.
For intensity normalization, Hounsfield Units are clipped to the range $[-1000, 1000]$ and linearly scaled to $[0,1]$.  
After preprocessing, we generate 50 projection images at a resolution of $512 \times 512$ for each volume using the TIGRE toolbox~\cite{biguri2016tigre}.

\section{Additional Quantitative comparison}
\cref{tab:8v} presents an additional quantitative comparison with optimization-based methods in the 8-view setting. Our method achieves 33.35 PSNR and 0.919 SSIM, outperforming all baselines. The baseline, R\textsuperscript{2}-Gaussian~\cite{zha2024r}, attains 26.08 PSNR and 0.805 SSIM, resulting in a substantial 7.27\,dB PSNR gap.
These results, together with the main experiments in~\cref{sec:experiment}, demonstrate that \textit{ILV} preserves high reconstruction quality even under extremely sparse projection settings and provides clear advantages over optimization-based approaches in such scenarios.

\section{Fine-tuning Details}

\begin{wrapfigure}{r}{0.4\linewidth}
\vspace{-2\baselineskip}
  \centering
  \includegraphics[width=\linewidth]{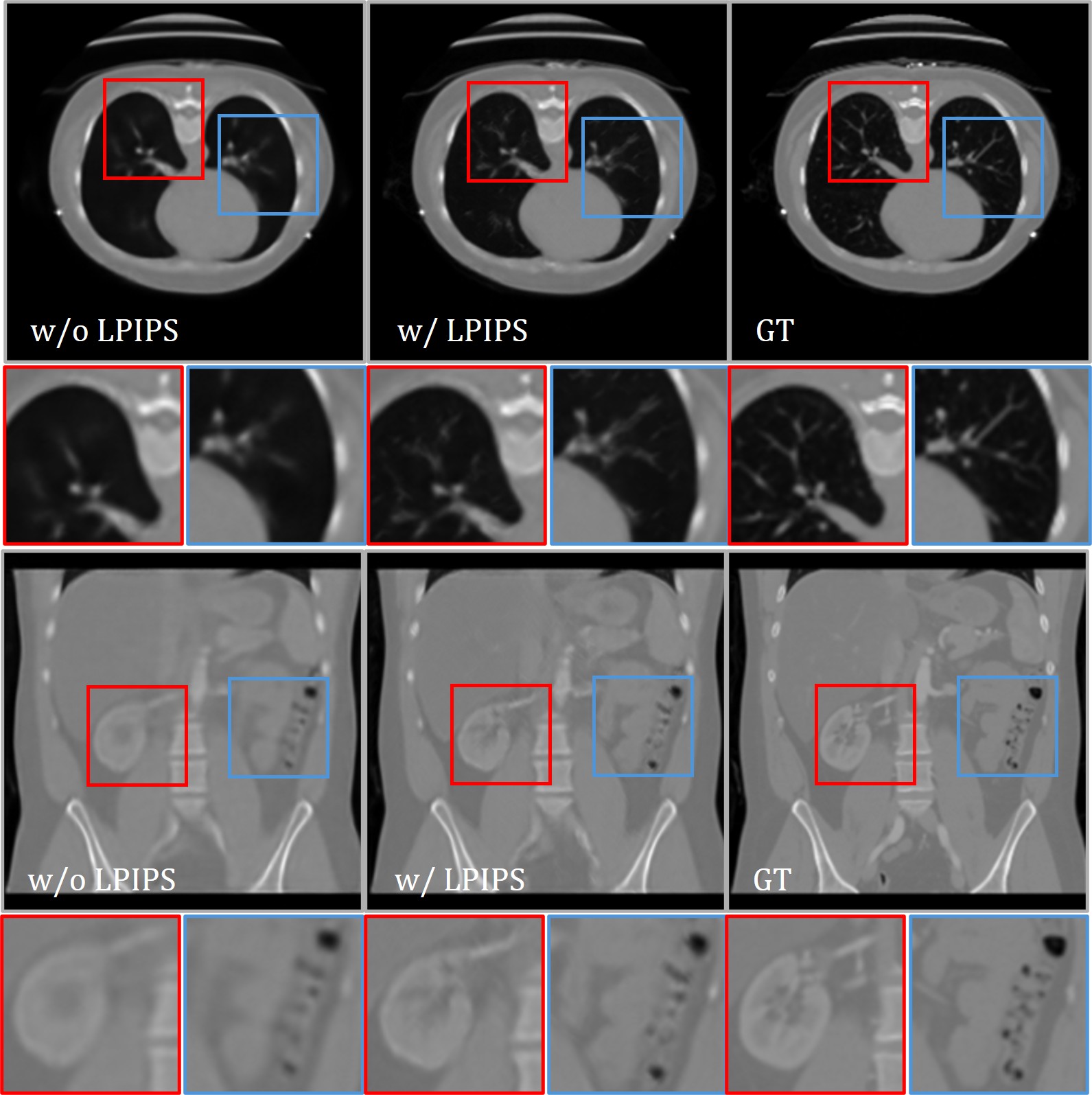}
  \vspace{-0.6cm}
  \caption{Qualitative results of perceptual fine-tuning.}
  \label{supple_lpips}
\end{wrapfigure}

In this section, we provide additional fine-tuning details and further qualitative results. As discussed in~\cref{sec:finetuning} applying an MSE or MAE loss directly to the reconstructed volume tends to produce overly smooth and blurry CT results. To mitigate this issue, we incorporate a perceptual constraint by sampling 20\% of 2D slices across all anatomical axes and computing the LPIPS loss, which is averaged over the sampled slices and added to the overall objective. Due to memory limitations, we exclude the rendering loss during fine-tuning.

We fine-tune the model for 10 epochs with an initial learning rate of $2\times10^{-5}$ while keeping all other settings identical to those used in the main experimental configuration. As shown in~\cref{supple_lpips}, introducing the LPIPS loss during fine-tuning leads to noticeably sharper and more detailed reconstructions.

\begin{table}[h]
\centering

\begin{minipage}[t]{0.48\linewidth}
\centering
\scriptsize
\caption{Quantitative comparison on novel-view X-ray synthesis in the 8-view setting.}
\setlength{\tabcolsep}{10pt}
\renewcommand{\arraystretch}{1.1}
\vspace{-1\baselineskip}
\begin{tabular}{ccc}
\toprule[0.1em]
Method & PSNR$\uparrow$ & SSIM$\uparrow$ \\
\midrule
IntraTomo~\cite{zang2021intratomo} & 33.94 & 0.987\\
NAF~\cite{zhaneural} & 33.37 & 0.917 \\
SAX-NeRF~\cite{cai2024structure} & 36.64 & \textbf{0.993} \\
R$^2$-Gaussian~\cite{zha2024r} & 37.88 & 0.980 \\
Ours & \textbf{43.33} & 0.992 \\
\bottomrule[0.1em]
\end{tabular}
\label{tab:nvs-8v}
\end{minipage}
\hfill
\begin{minipage}[t]{0.48\linewidth}
\centering
\scriptsize
\caption{Quantitative comparison on novel-view X-ray synthesis in the 10-view setting.}
\setlength{\tabcolsep}{10pt}
\renewcommand{\arraystretch}{1.1}
\vspace{-1\baselineskip}
\begin{tabular}{ccc}
\toprule[0.1em]
Method & PSNR$\uparrow$ & SSIM$\uparrow$ \\
\midrule
IntraTomo~\cite{zang2021intratomo} & 35.03 & 0.990\\
NAF~\cite{zhaneural} & 34.78 & 0.929 \\
SAX-NeRF~\cite{cai2024structure} & 38.44 & \textbf{0.995} \\
R$^2$-Gaussian~\cite{zha2024r} & 39.97 & 0.984 \\
Ours & \textbf{44.20} & 0.993 \\
\bottomrule[0.1em]
\end{tabular}
\label{tab:nvs-10v}
\end{minipage}
\vspace{-3\baselineskip}
\end{table}

\begin{figure}[h]
  \centering
  \includegraphics[width=1\linewidth]{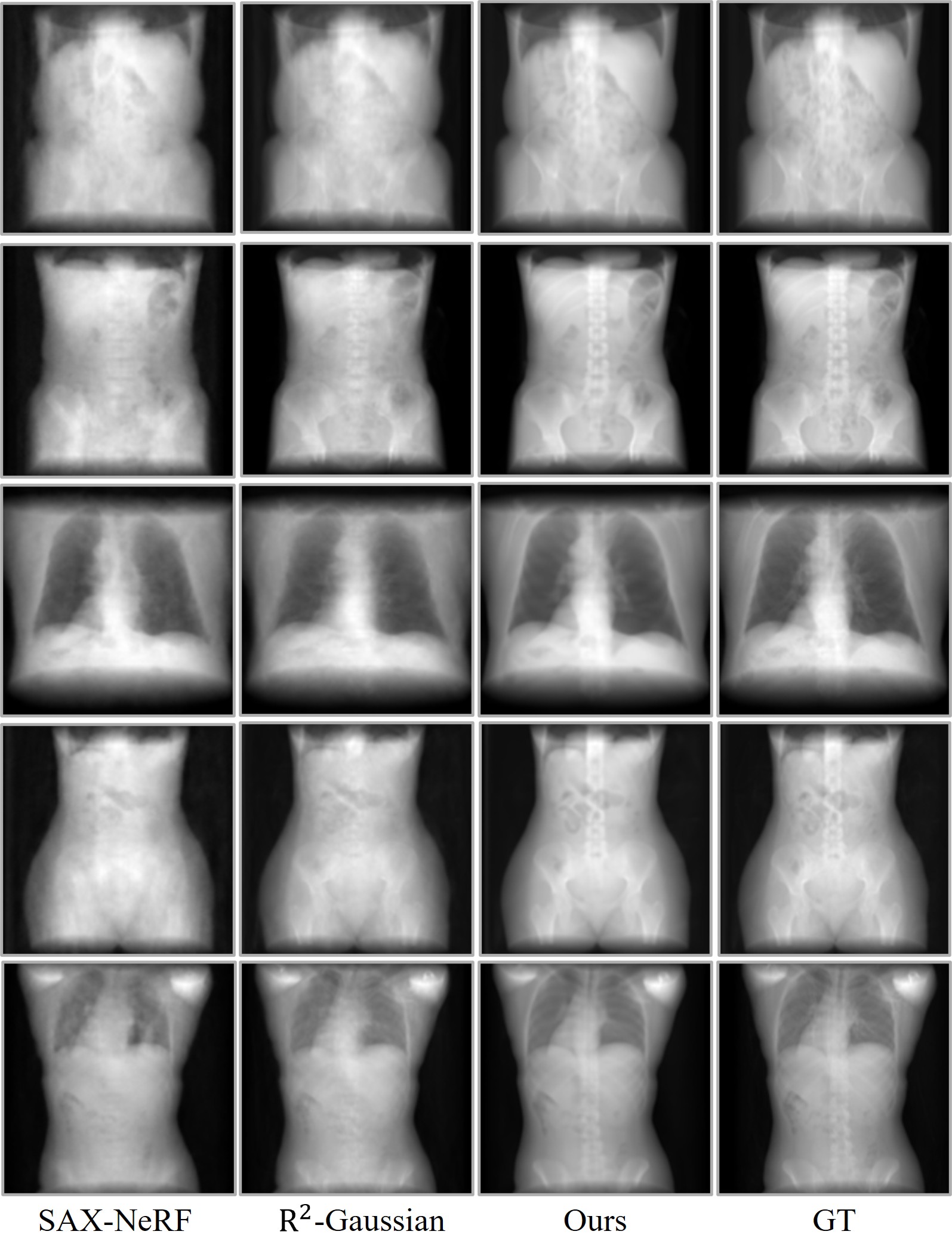}
  \caption{Qualitative comparison on novel view X-ray synthesis in the 10-view setting.}
  \label{supple_nvs}
\end{figure}

\section{Additional Novel View X-ray Synthesis}
In this section, we provide additional comparisons of novel-view X-ray synthesis performance under the 8-view and 10-view settings.~\cref{tab:nvs-8v} and \cref{tab:nvs-10v} summarize the quantitative results for each setting using various optimization-based baselines. In the 8-view setting, our method achieves an improvement of approximately 5.45\,dB over R$^2$-Gaussian~\cite{zha2024r}.
In the 10-view setting, our method also outperforms R$^2$-Gaussian by about 4.23\,dB in PSNR.
~\cref{supple_nvs} presents qualitative comparisons in the 10-view setting. SAX-NeRF~\cite{cai2024structure} and R$^2$-Gaussian fail to accurately recover fine anatomical structures such as bone details, whereas our method preserves these high-frequency structures more effectively.

\section{Additional Qualitative results}

In this section, we provide additional qualitative comparisons to complement the main paper.
\cref{supple1,supple2,supple3,supple4} show the reconstruction results in the 10-view scenario. Traditional and feed-forward approaches such as SART~\cite{andersen1984simultaneous} and DIF-Net~\cite{lin2023learning} fail to recover meaningful anatomical structures, exhibiting strong artifacts or heavy blurring that leads to significant loss of detail. The 2D slice-based method FreeSeed~\cite{ma2023freeseed} shows limited consistency across slices. In the coronal and sagittal views, this appears as a layered or slightly unstable pattern, which is caused by processing each slice independently. DIF-Gaussian~\cite{lin2024learning} recovers only coarse shapes. Bony regions appear faint, and soft tissues and internal organs are mostly missing or only weakly visible. In comparison, \textit{ILV} provides clearer bony structures and more stable soft-tissue reconstruction across all anatomical planes.

\cref{supple6,supple7,supple8,supple9,supple10} present comparisons with optimization-based methods in the 24-view setting. IntraTomo~\cite{zang2021intratomo} produces heavily blurred results and fails to preserve most internal structures. NAF~\cite{zhaneural} and SAX-NeRF~\cite{cai2024structure} produce noisy reconstructions, making it difficult to clearly identify anatomical boundaries.
R$^2$-Gaussian~\cite{zha2024r} recovers some structural details more effectively, but the results contain the characteristic spiky artifacts commonly observed in Gaussian-based representations. In contrast, \textit{ILV} provides more detailed reconstructions and maintains clearer and more stable structures compared to existing optimization-based approaches.

\section{Qualitative Results for the Ablation Study}

\begin{wrapfigure}{r}{0.45\linewidth}
 \vspace{-0.8cm}
  \centering
  \includegraphics[width=\linewidth]{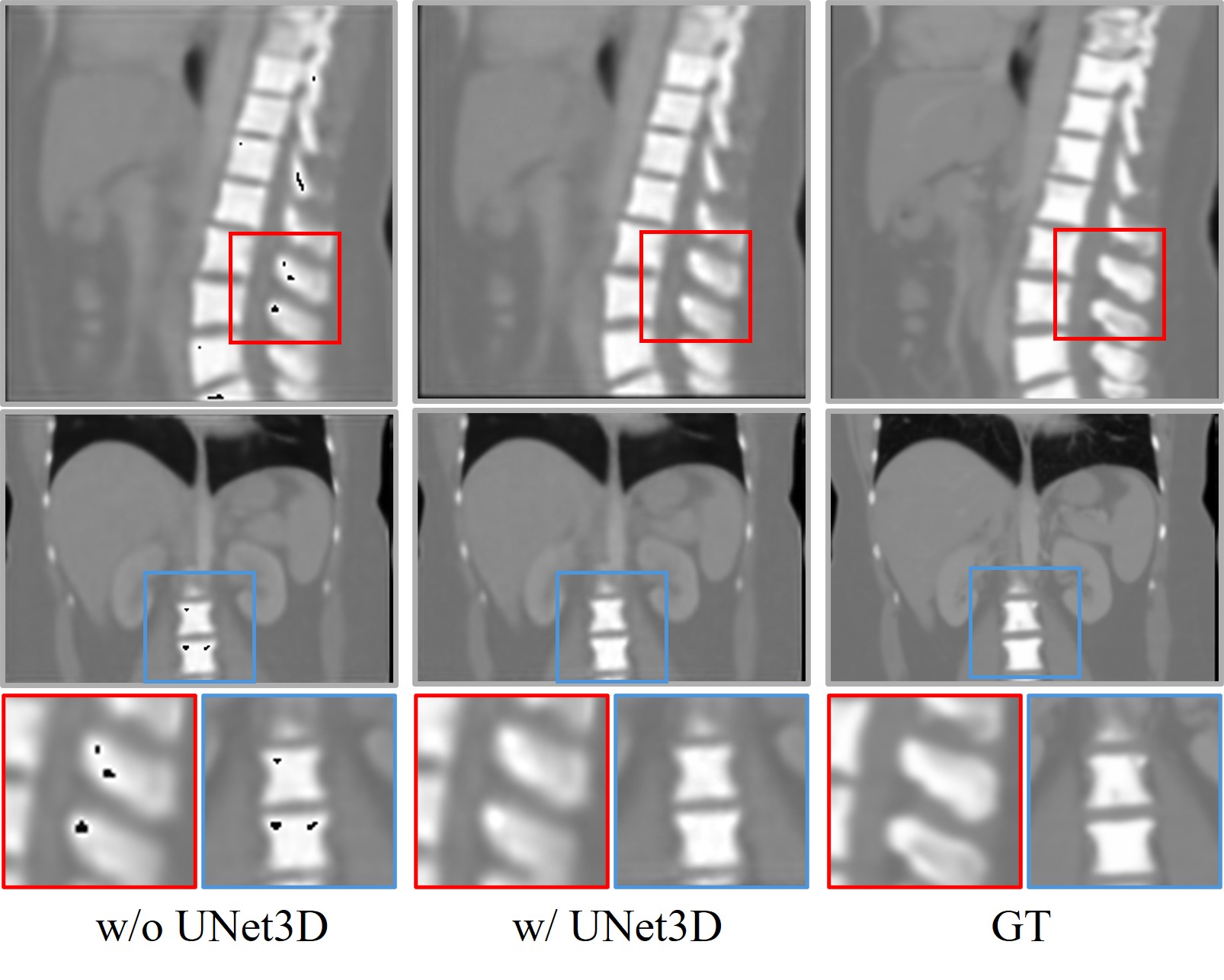}
   \vspace{-0.8cm}
  \caption{Qualitative results for the ablation study.}
  \label{supple_ab}
\end{wrapfigure}

In this section, we provide additional visual results that support the ablation study.
\cref{supple_ab} illustrates the behavior of the 3D U-Net refinement module, which is activated only after a certain predefined number of training iterations. Before the refinement stage is enabled, the coarse CT volume often exhibits broken black voxel regions, indicating unstable intensity behavior. Once the refinement module becomes active, these artifacts disappear and the reconstruction becomes structurally stable and consistent.

\begin{figure*}[t]
  \centering
  \includegraphics[width=1\textwidth]{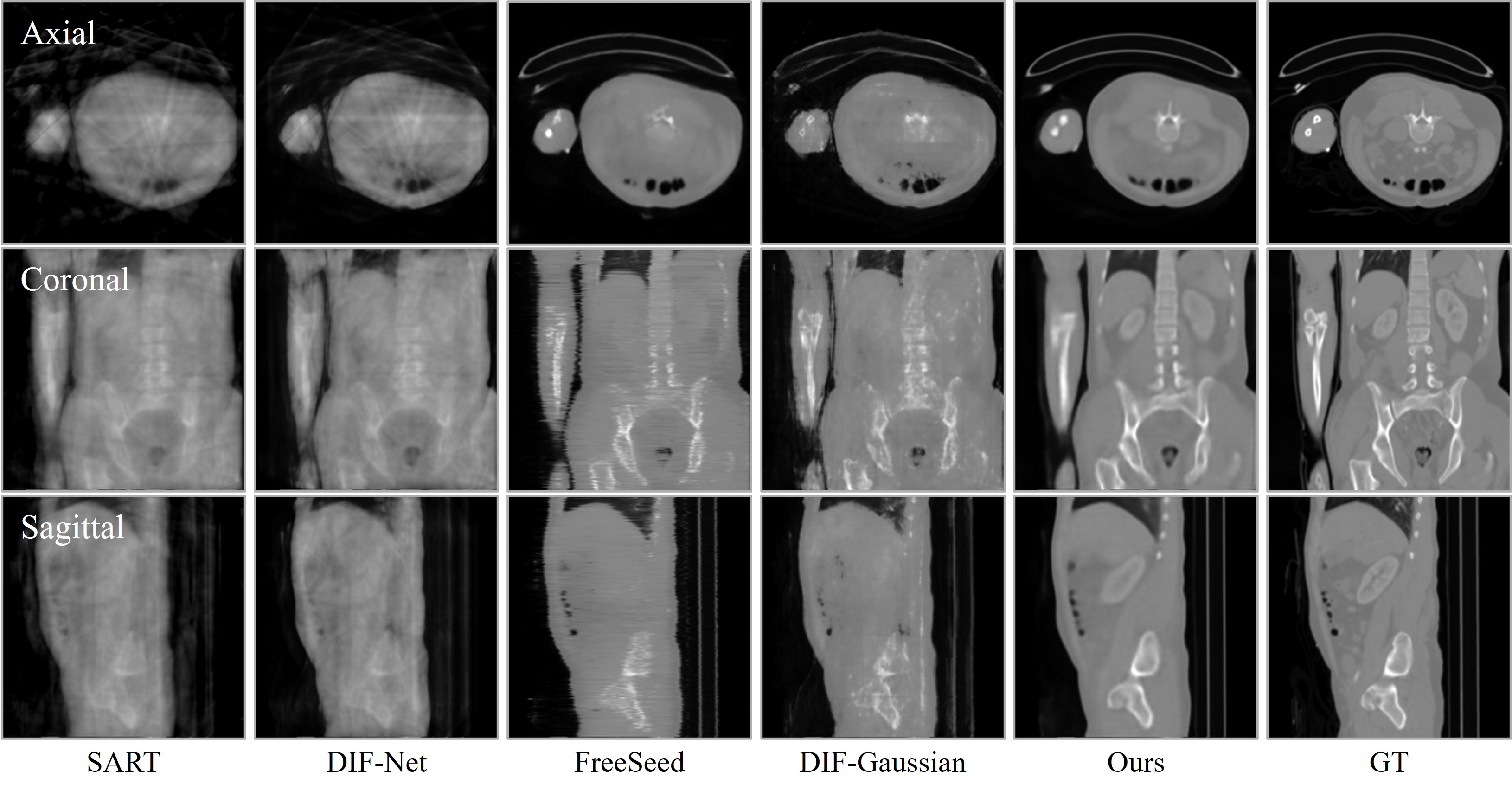}
  \caption{Qualitative comparison with traditional and feed-forward methods on CT reconstruction in the 10-view setting.}
  \label{supple1}
\end{figure*}

\begin{figure*}[t]
  \centering
  \includegraphics[width=1\textwidth]{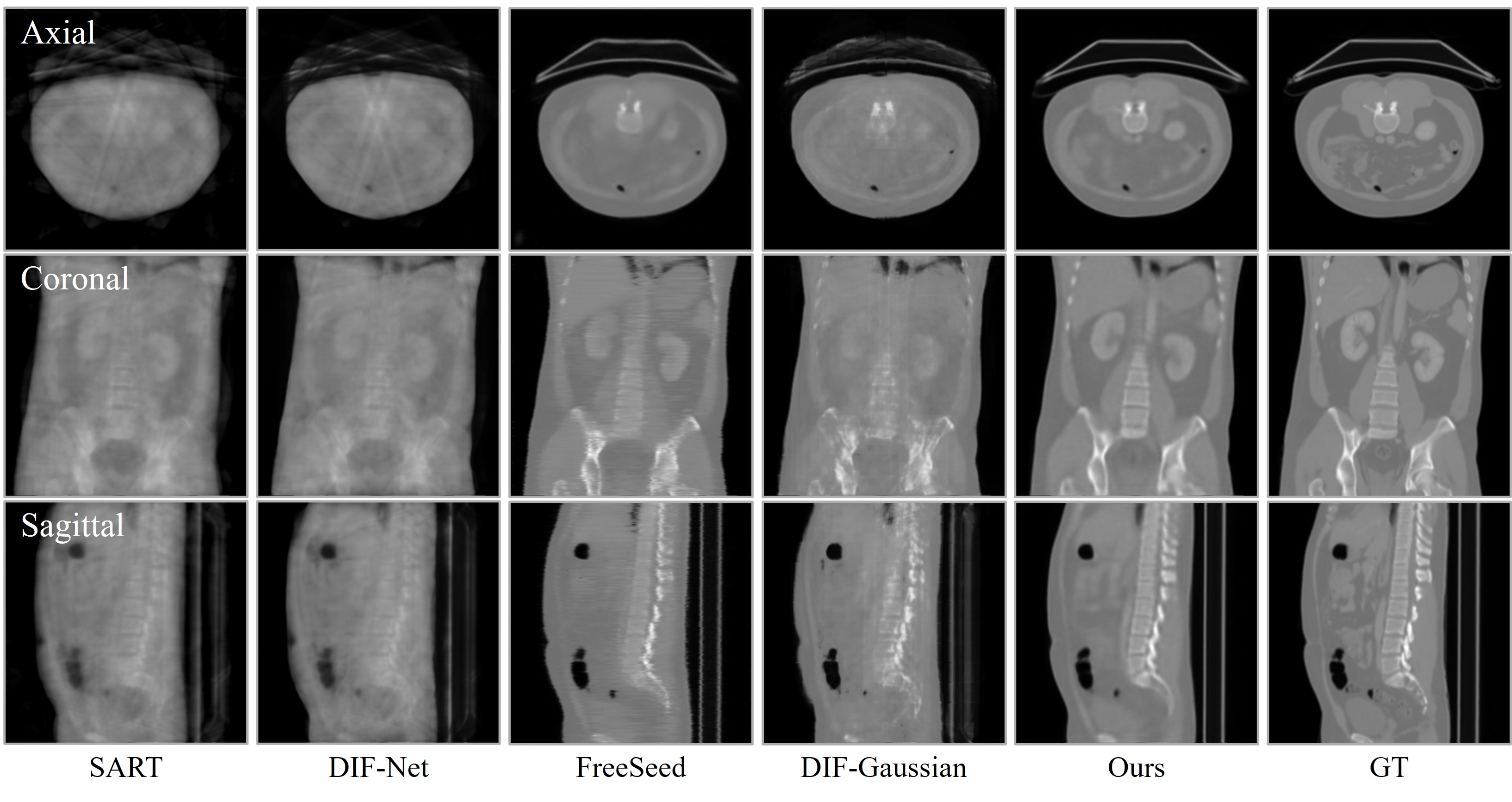}
  \caption{Qualitative comparison with traditional and feed-forward methods on CT reconstruction in the 10-view setting.}
  \label{supple2}
\end{figure*}

\begin{figure*}[t]
  \centering
  \includegraphics[width=1\textwidth]{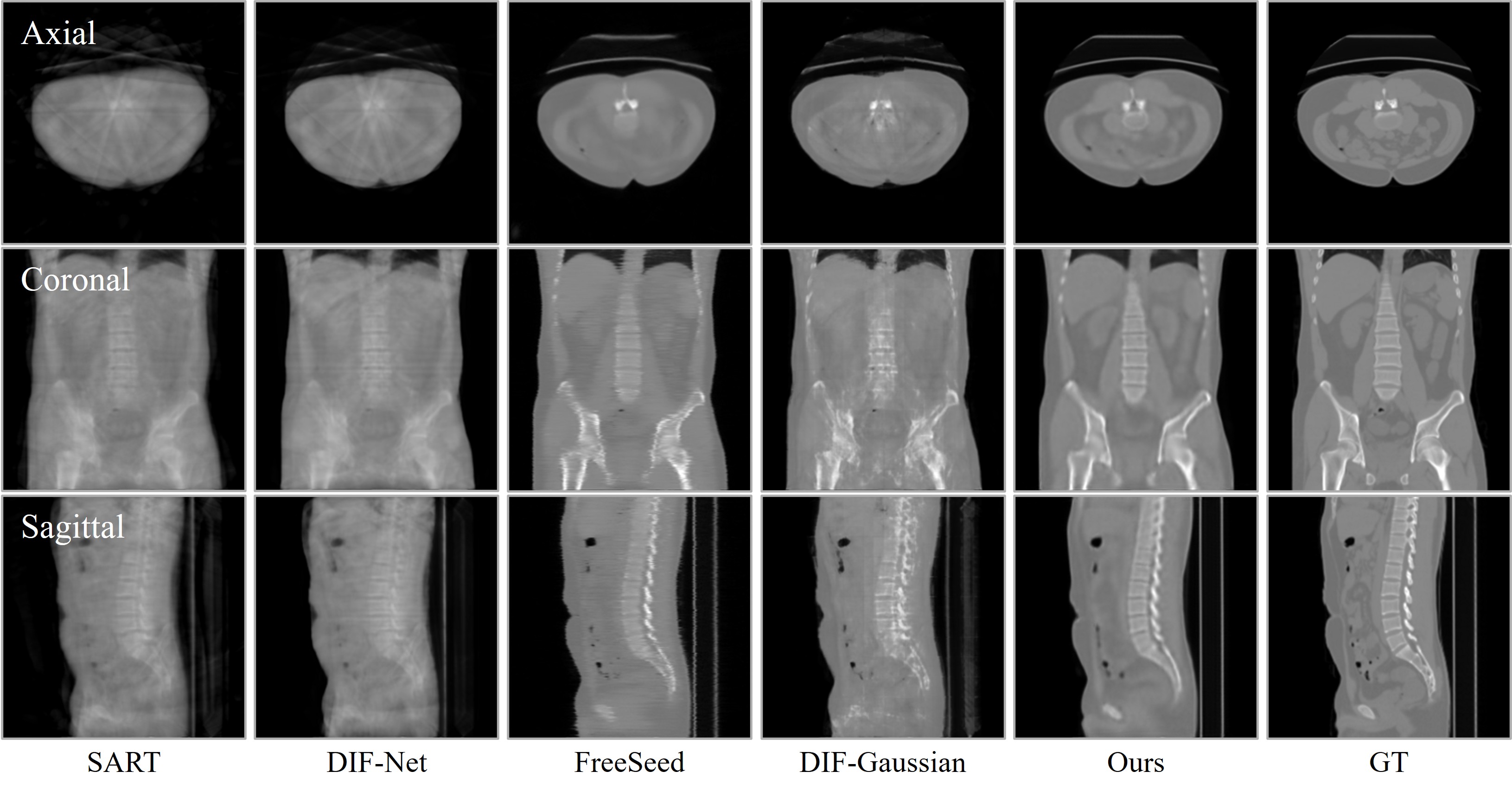}
  \caption{Qualitative comparison with traditional and feed-forward methods on CT reconstruction in the 10-view setting.}
  \label{supple3}
\end{figure*}

\begin{figure*}[t]
  \centering
  \includegraphics[width=1\textwidth]{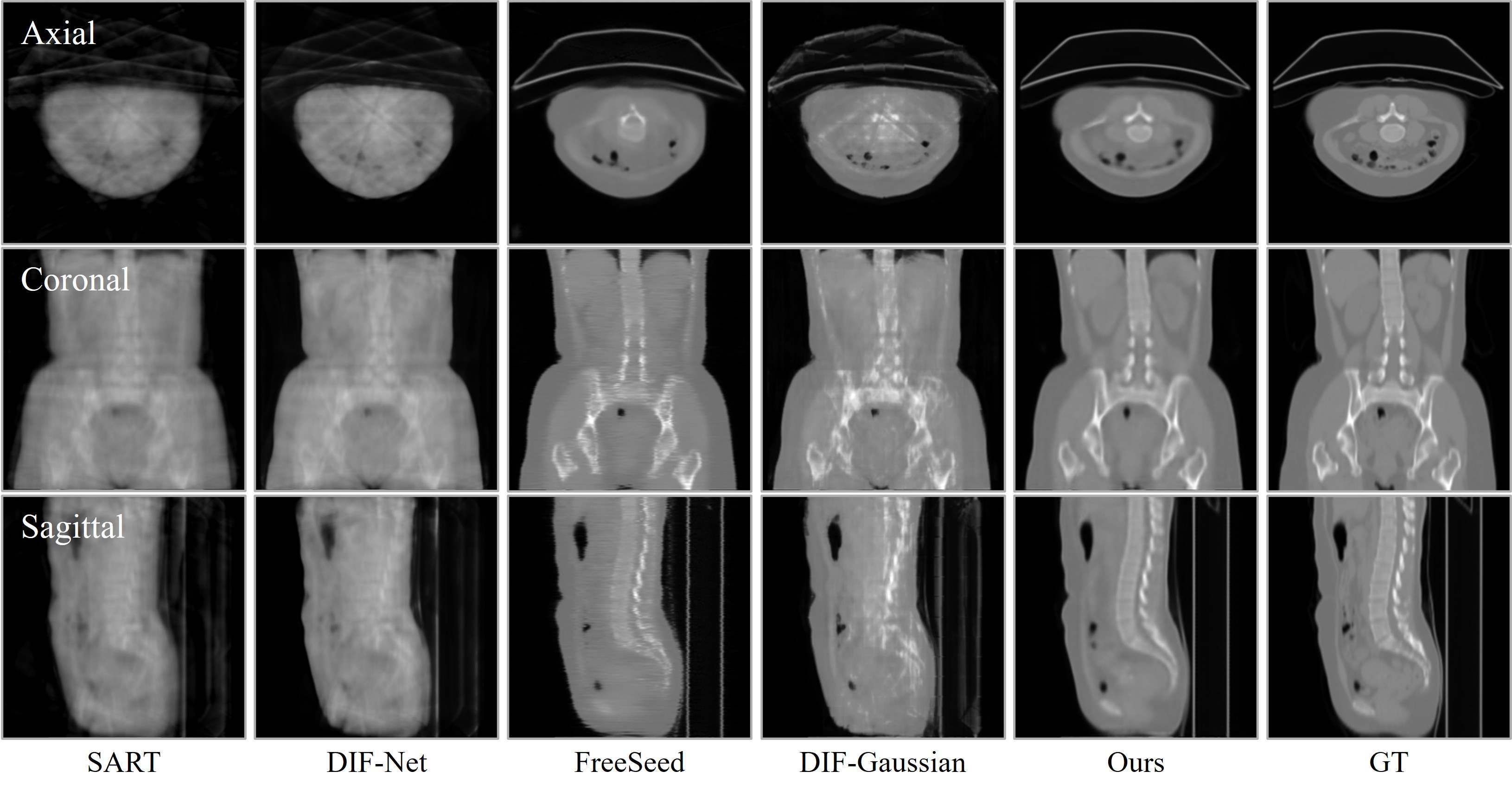}
  \caption{Qualitative comparison with traditional and feed-forward methods on CT reconstruction in the 10-view setting.}
  \label{supple4}
\end{figure*}

\begin{figure*}[t]
  \centering
  \includegraphics[width=1\textwidth]{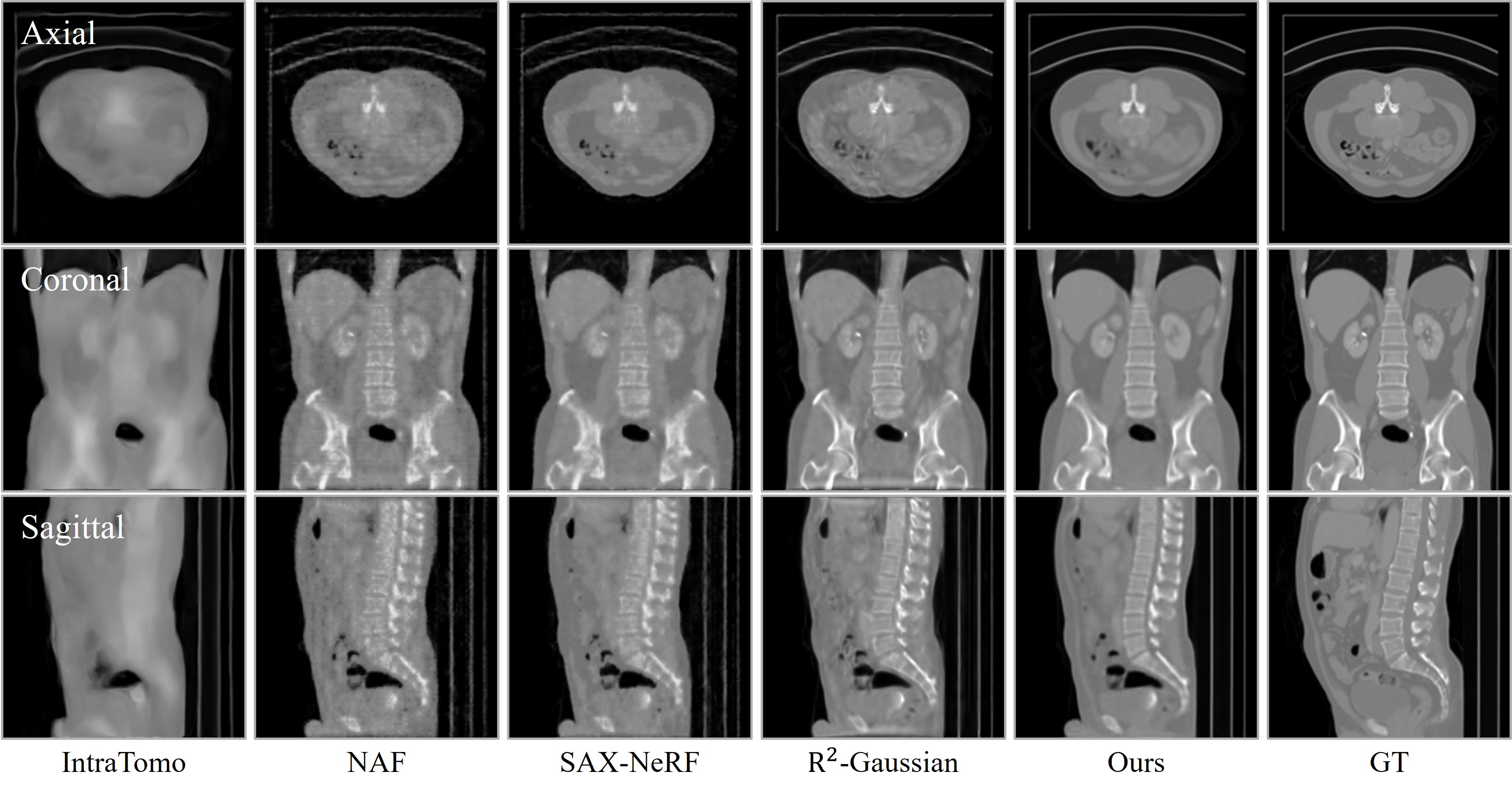} 
  \caption{Qualitative comparison with optimization-based methods on CT reconstruction in the 24-view setting.}
  \label{supple5}
\end{figure*}

\begin{figure*}[t]
  \centering
  \includegraphics[width=1\textwidth]{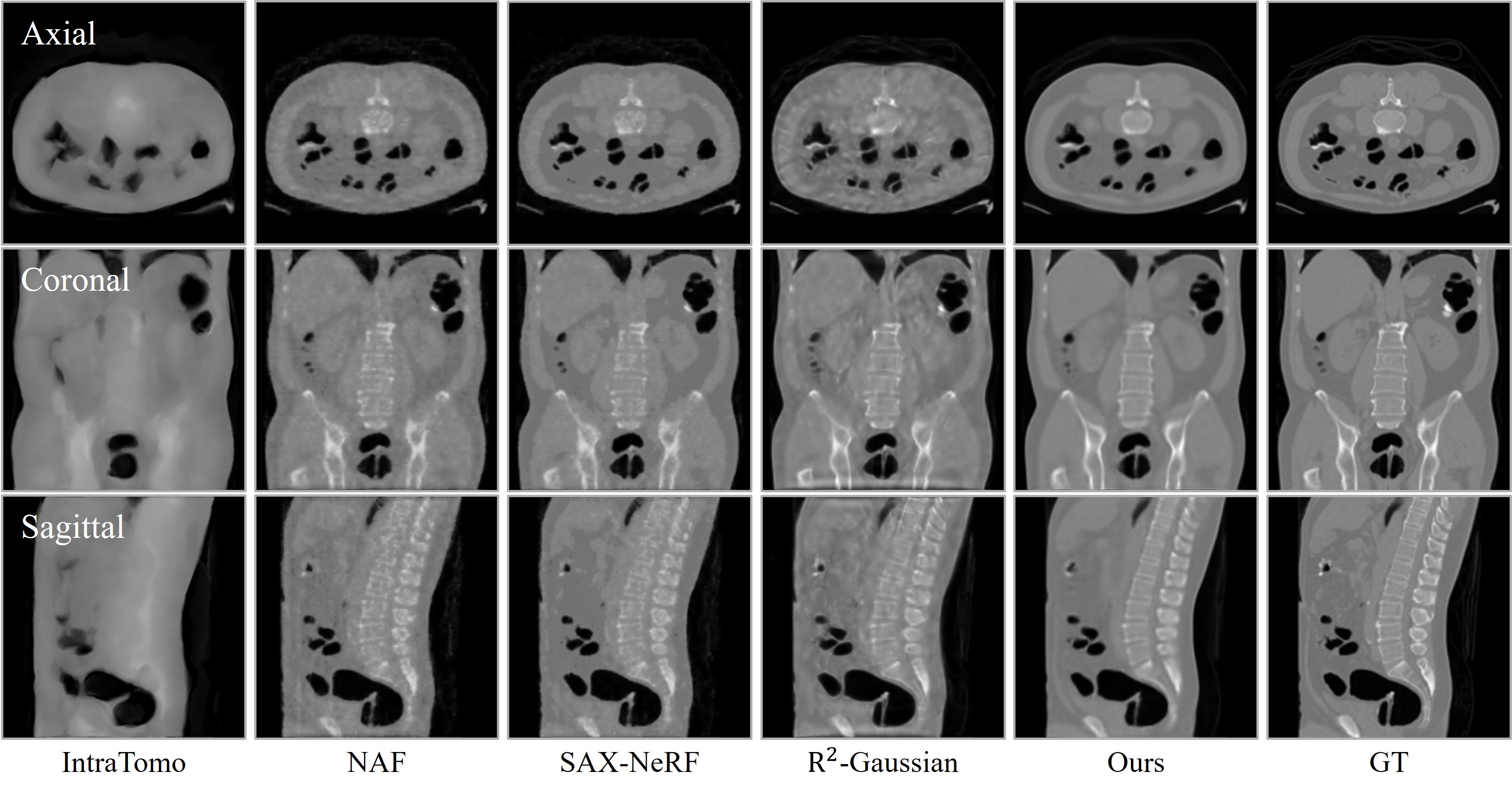} 
  \caption{Qualitative comparison with optimization-based methods on CT reconstruction in the 24-view setting.}
  \label{supple6}
\end{figure*}

\begin{figure*}[t]
  \centering
  \includegraphics[width=1\textwidth]{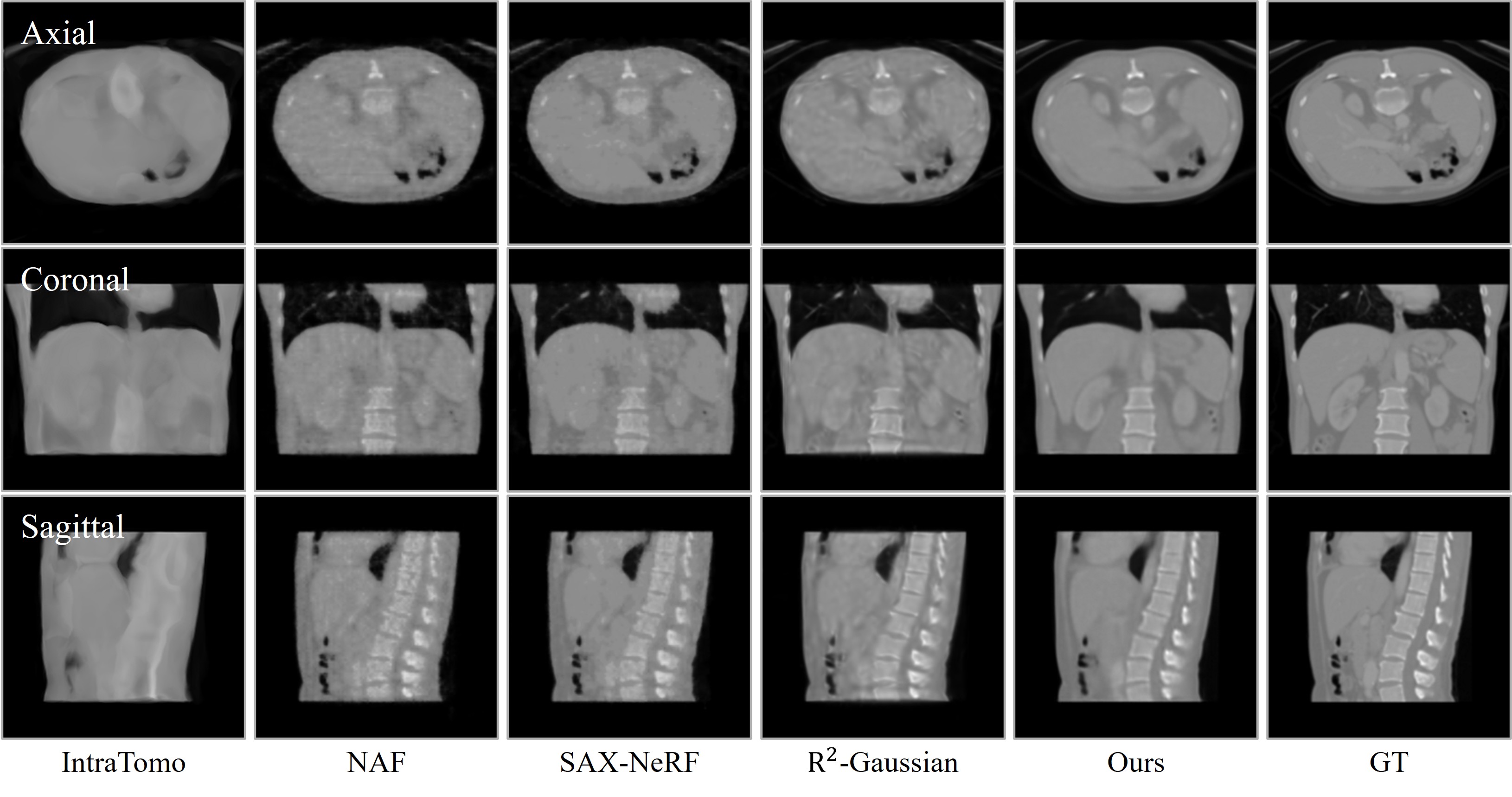} 
  \caption{Qualitative comparison with optimization-based methods on CT reconstruction in the 24-view setting.}
  \label{supple7}
\end{figure*}

\begin{figure*}[t]
  \centering
  \includegraphics[width=1\textwidth]{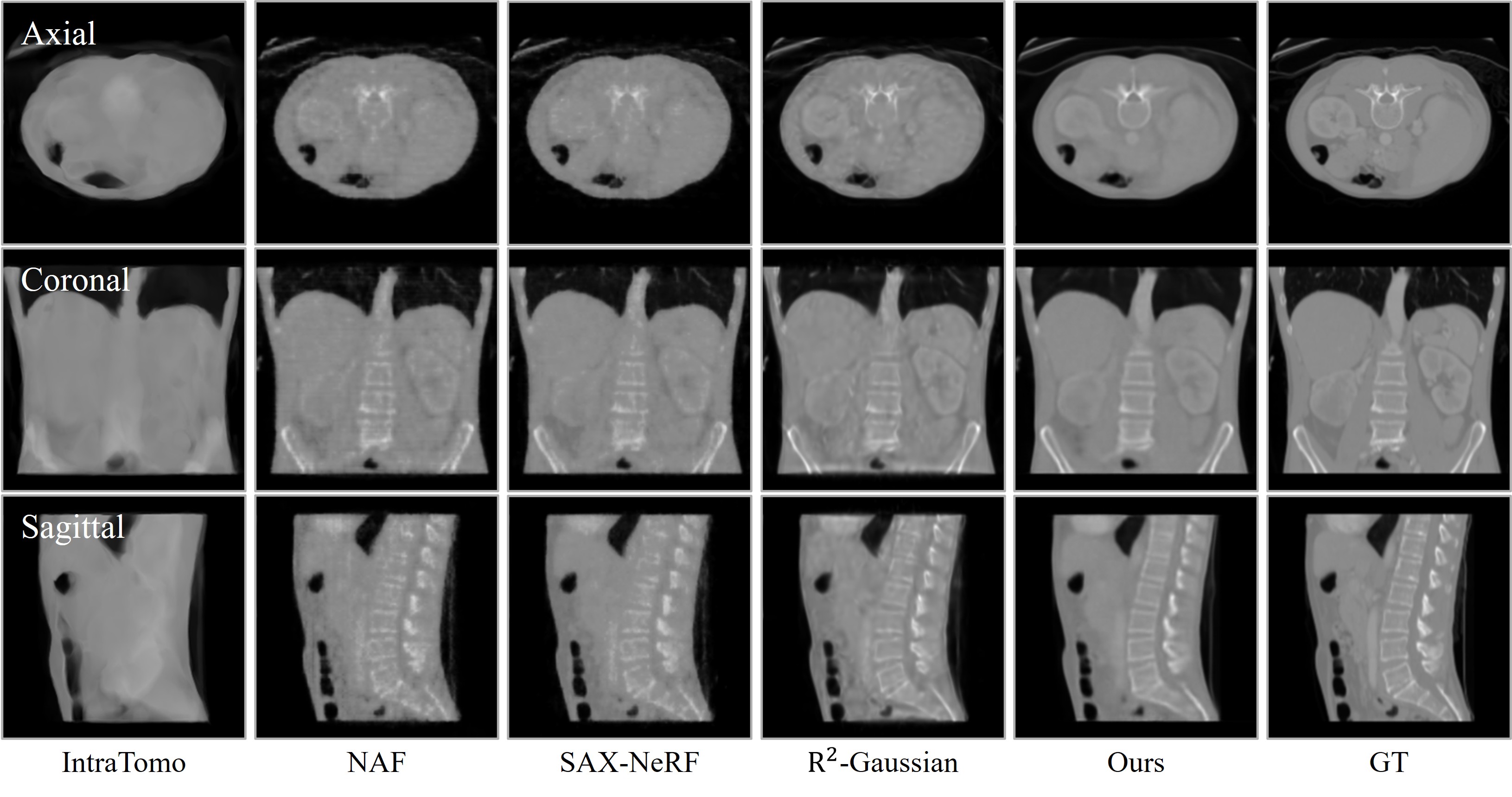} 
  \caption{Qualitative comparison with optimization-based methods on CT reconstruction in the 24-view setting.}
  \label{supple8}
\end{figure*}

\begin{figure*}[t]
  \centering
  \includegraphics[width=1\textwidth]{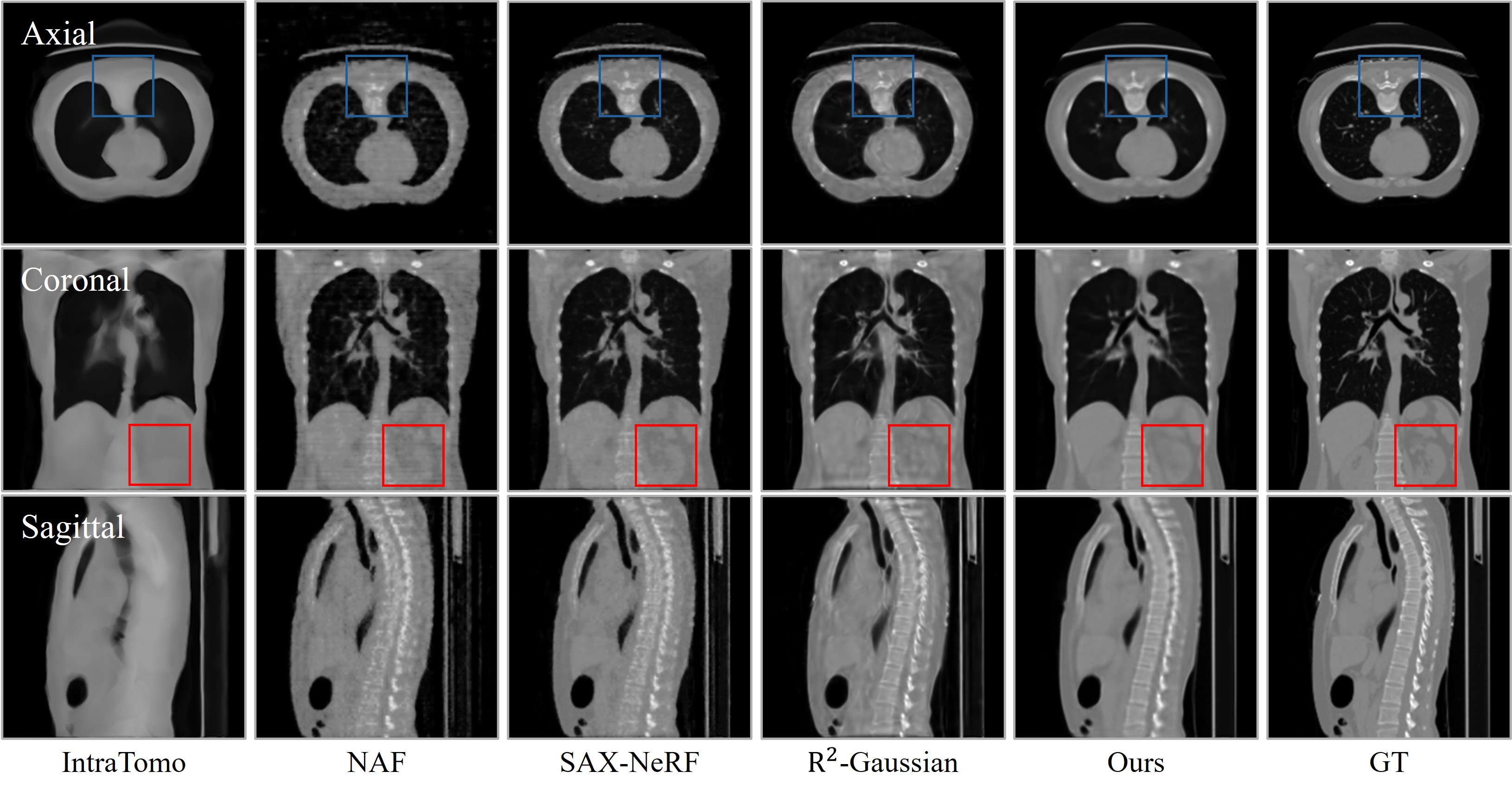} 
  \caption{Qualitative comparison with optimization-based methods on CT reconstruction in the 24-view setting.}
  \label{supple9}
\end{figure*}

\begin{figure*}[t]
  \centering
  \includegraphics[width=1\textwidth]{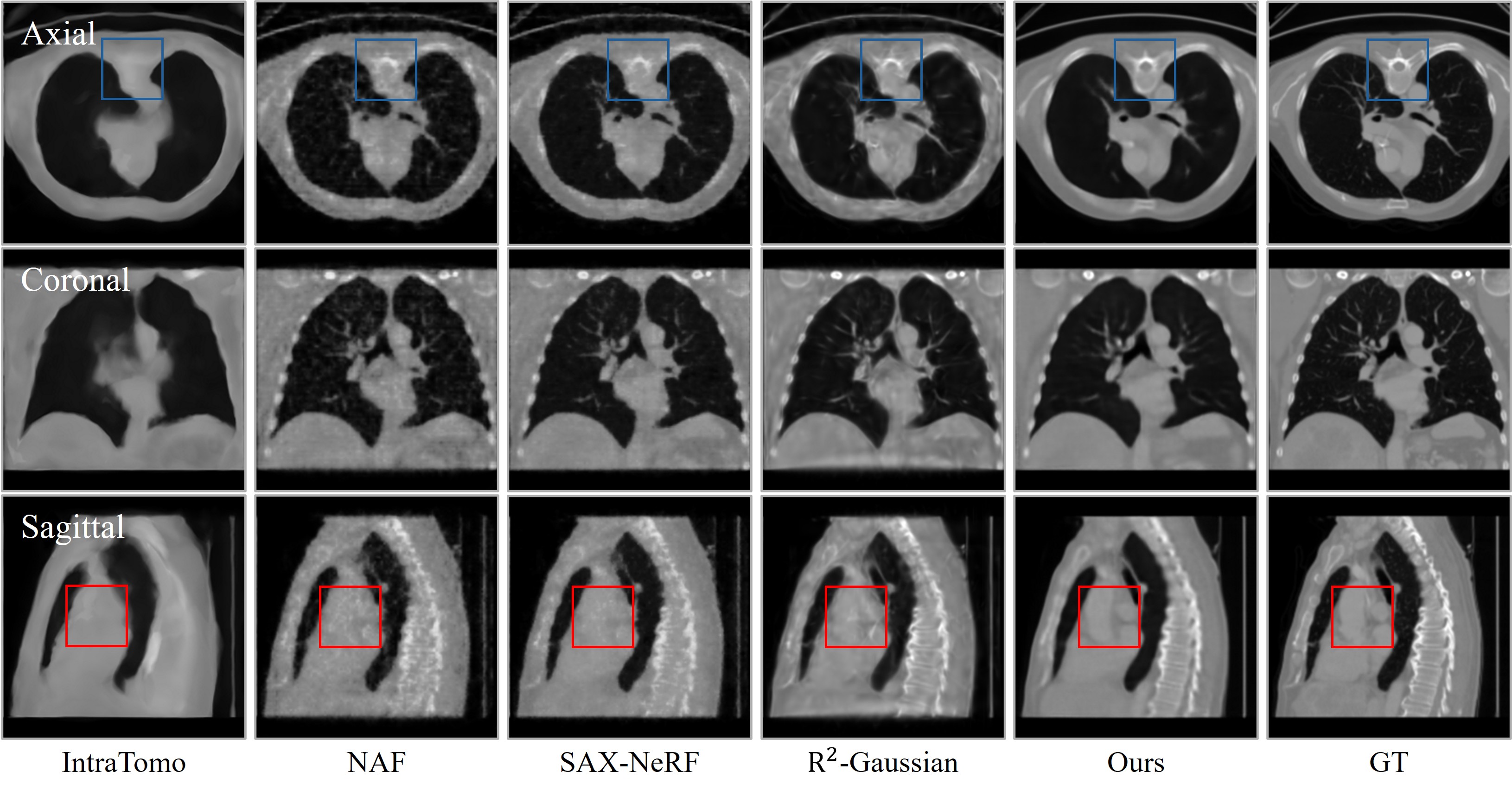} 
  \caption{Qualitative comparison with optimization-based methods on CT reconstruction in the 24-view setting.}
  \label{supple10}
\end{figure*}

\end{document}